\documentclass[journal]{IEEEtai}

\usepackage[colorlinks,urlcolor=blue,linkcolor=blue,citecolor=blue]{hyperref}

\usepackage{epstopdf} 
\usepackage{color,array}
\usepackage{graphicx}
\usepackage{comment}
\usepackage{booktabs}
\usepackage{multirow}
\usepackage{pifont}
\usepackage{bbding}
\usepackage{float}
\usepackage{amsfonts,amsthm,amsmath,amssymb}
\usepackage{subfigure}
\usepackage{xurl}
\usepackage{xparse}
\usepackage{flushend}
\usepackage{wrapfig}
\usepackage{xfrac}
\usepackage{cleveref}
\usepackage[table,xcdraw]{xcolor}

\newcommand{\highlight}{0}

\ifnum\highlight=1
  \newcommand{\red}[1]{{\color[HTML]{FF0000} #1}}
\else
  \newcommand{\red}[1]{#1}
\fi

\ifnum\highlight=2
    \newcommand{\revise}[1]{{\color[HTML]{FF0000} #1}}
\else
  \newcommand{\revise}[1]{#1}
\fi

\ifnum\highlight=3
    \newcommand{\revision}[1]{{\color[HTML]{FF0000} #1}}
\else
  \newcommand{\revision}[1]{#1}
\fi

\ifnum\highlight=4
    \newcommand{\modify}[1]{{\color[HTML]{FF0000} #1}}
\else
  \newcommand{\modify}[1]{#1}
\fi

\ifnum\highlight=5
    \newcommand{\rfive}[1]{{\color[HTML]{FF0000} #1}}
\else
  \newcommand{\rfive}[1]{#1}
\fi

\ifnum\highlight=6
    \newcommand{\rsix}[1]{{\color[HTML]{FF0000} #1}}
\else
  \newcommand{\rsix}[1]{#1}
\fi

\ifnum\highlight=7
    \newcommand{\rfinal}[1]{{\color[HTML]{FF0000} #1}}
\else
  \newcommand{\rfinal}[1]{#1}
\fi

\newcommand{\redbf}[1]{{\color[HTML]{FF0000} \textbf{#1}}}
\newcommand{\dualline}[1]{\begin{tabular}[c]{@{}l@{}}#1\end{tabular}}
\newcommand{\vc}[1]{\multicolumn{1}{c}{\multirow{-2}{*}{#1}}}

\begin{document}

\title{WeatherRemover: All-in-one Adverse Weather Removal with Multi-scale Feature Map Compression} 

\author{Weikai Qu, Sijun Liang, Cheng Pan, Zikuan Yang, Guanchi Zhou, Xianjun Fu, \\Bo Liu, Changmiao Wang*, and Ahmed Elazab
\thanks{This work was supported by GuangDong Basic and Applied Basic Research Foundation (No. 2025A1515011617), National Key Research and Development Program of China (No. 2025YFE0103200), Guangxi Science and Technology Program (No. FN2504240022), Shenzhen Medical Research Fund (No. C2401036), and the Project (No. 20232ABC03A25).
\textit{Corresponding author: Changmiao Wang}}
\thanks{Weikai Qu, Guangdong University of Technology (e-mail: weikaiqu57@gmail.com)}
\thanks{Sijun Liang, Guangdong University of Technology (e-mail: chrisliang97@gmail.com)}
\thanks{Cheng Pan, Sanda University (e-mail: panc@sandau.edu.cn)}
\thanks{Zikuan Yang, University of Queensland (e-mail: s4752340@uq.edu.au)}
\thanks{Guanchi Zhou, Shenzhen MSU-BIT University (e-mail: zhouguanchi437@gmail.com)}
\thanks{Xianjun Fu, Zhejiang college of Security Technology (e-mail: fuxianjun@zjcst.edu.cn)}
\thanks{Bo Liu, Northwest China Research Institute of Electronic Equipment (e-mail: liubo\_nwiee@163.com)}
\thanks{Changmiao Wang, Shenzhen Research Institute of Big Data (e-mail: cmwangalbert@gmail.com)}
\thanks{Ahmed Elazab, Shenzhen University and Misr Higher Institute for Commerce and Computers (e-mail: ahmedelazab@szu.edu.cn)}}

\markboth{IEEE TRANSACTIONS ON ARTIFICIAL INTELLIGENCE, Vol. 00, No. 0, November 2026}
{QU \MakeLowercase{\textit{et al.}}: WeatherRemover: All-in-one Adverse Weather Removal with Multi-scale Feature Map Compression}

\maketitle

\begin{abstract}
\rsix{Photographs taken in adverse weather conditions often suffer from blurriness, occlusion, and low brightness due to interference from rain, snow, and fog. These issues can significantly hinder the performance of subsequent computer vision tasks, making the removal of weather effects a crucial step in image enhancement.} Existing methods primarily target specific weather conditions, with only a few capable of handling multiple weather scenarios. \red{However, mainstream approaches often overlook performance considerations, resulting in large parameter sizes, long inference times, and high memory costs.} In this study, we introduce the WeatherRemover model, \red{designed to enhance the restoration of images affected by various weather conditions while balancing performance.} \rsix{Our model adopts a UNet-like structure with a gating mechanism and a multi-scale pyramid vision Transformer. It employs channel-wise attention derived from convolutional neural networks to optimize feature extraction, while linear spatial reduction helps curtail the computational demands of attention. The gating mechanisms, strategically placed within the feed-forward and downsampling phases, refine the processing of information by selectively addressing redundancy and mitigating its influence on learning. This approach facilitates the adaptive selection of essential data, ensuring superior restoration and maximizing efficiency.} \modify{Additionally, our lightweight model achieves an optimal balance between restoration quality, parameter efficiency, computational overhead, and memory usage, distinguishing it from other multi-weather models, thereby meeting practical application demands effectively.} The source code is available at \url{https://github.com/RICKand-MORTY/WeatherRemover}.
\end{abstract}

\begin{IEEEImpStatement}
 Removing the interference caused by adverse weather conditions in images is crucial in image enhancement. However, most current models are designed to handle only single-weather scenarios. A few models can address multiple weather conditions, but they often suffer from inefficiencies due to large parameter sizes, long inference times, or suboptimal restoration effects. The WeatherRemover model proposed in this paper can efficiently handle single-weather and multi-weather tasks. Experimental results demonstrate that our model excels in single and multi-weather removal tasks. Moreover, the model's relatively lightweight configuration underscores its efficiency and flexibility. 
\end{IEEEImpStatement}


\section{Introduction}

\begin{figure*}[!t]
  \centering
  \includegraphics[scale=0.5]{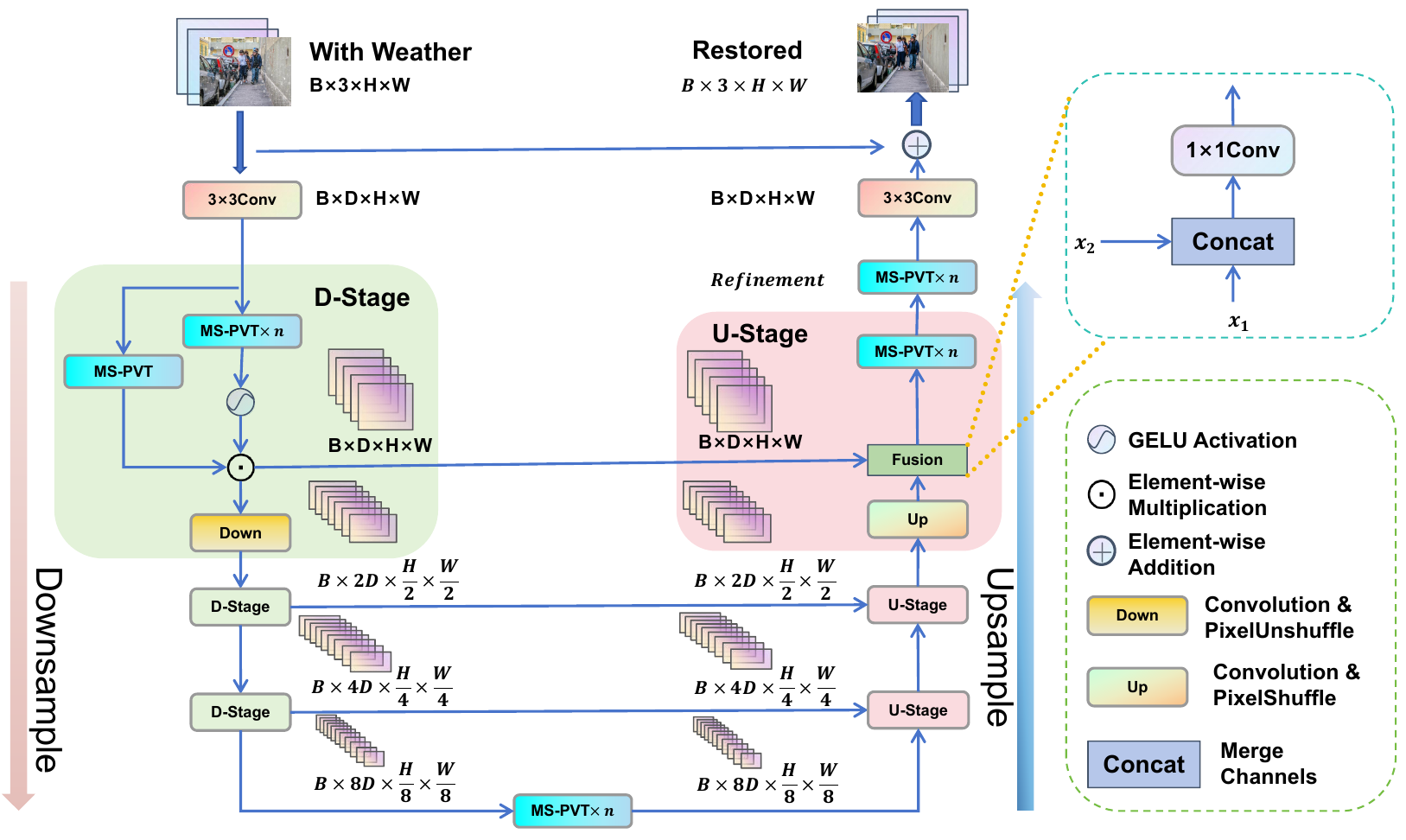}
  \caption{\revise{Overview of WeatherRemover structure.} The WeatherRemover model is structured around a UNet architecture. \revise{The left side of the model is dedicated to downsample, incorporating a gating mechanism to manage the flow of information, delineated as three D-Stages. \rfinal{The gating mechanism in D-Stage employs a dual-branch structure, where one branch is activated via the GELU function and both branches are fused through element-wise multiplication. The right side performs upsampling, merges features from matching downsampling layers, and includes three U-Stages. Each U-Stage fuses outputs from the previous layer and matching D-Stage along channels, then applies a 1$\times$1 convolution to halve the channels. Each stage utilizes MS-PVT, halving spatial size and doubling channels during downsampling, and reversing this during upsampling.}}}
  \label{fig:backbone}
\end{figure*}

\IEEEPARstart{M}{itigating} weather-related interference in images is a crucial research area in computer vision. \rsix{Enhancing the performance of visual systems in adverse weather conditions is essential because environmental factors like rain, snow, and fog can significantly impair image quality. These conditions diminish contrast and introduce blurriness, complicating subsequent image processing and analysis.} This degradation affects the effectiveness of various visual systems~\cite{transweather}, including autonomous driving and smart cameras. \rsix{In addition, images taken in challenging weather conditions frequently experience interference from multiple sources, such as the simultaneous presence of rain and fog during heavy rainfall.} Consequently, removing weather-related interference from images presents a substantial challenge.

Conventional image restoration methods typically rely on handcrafted priors and physical scene modeling, but these approaches often struggle in challenging weather conditions. With the advent of deep learning, convolutional neural network (CNN)-based methods have become more prevalent~\cite{focalnet}. These methods learn a mapping from degraded to restored images, demonstrating more effectiveness in handling weather-induced degradation. However, they are limited by their limited receptive field and suboptimal adaptability to input content~\cite{cnct}. This hinders their ability to address large-scale degradation, such as heavy rain or snow. Moreover, the diverse spatial distribution of weather elements across different images presents a significant challenge for CNNs in effectively removing these effects.

In recent years, Transformers have garnered significant attention for their outstanding ability to handle sequential data, especially in natural language processing tasks. The introduction of the \red{vision transformer} (ViT) marked the entry of Transformers into the field of computer vision, yielding promising results~\cite{vit}. However, the substantial computational and memory requirements of ViT make it challenging to apply to high-resolution tasks~\cite{swin,restormer,pvt}. To address these challenges, the \red{pyramid vision transformer} (PVT) was developed. PVT is based on a feature pyramid structure~\cite{pvt,pvtv2}. Unlike ViT, PVT’s attention mechanism employs \modify{spatial reduction attention (SRA)} to reduce the size of the generated keys and values. This allows the attention mechanism to concentrate more on local features, thereby improving accuracy and performance.

Subsequent weather removal methods incorporated Transformers but required multiple decoders to handle various weather types, which proved ineffective for multi-scene processing~\cite{allinone}. Addressing this limitation, Transweather employed a single encoder-decoder framework to tackle multi-scene tasks, achieving notable success~\cite{transweather}. Following this, Restormer used a UNet-based~\cite{unet} backbone with CNN-based attention mechanisms, demonstrating robust performance across several image enhancement datasets~\cite{restormer}. DRS\red{f}ormer further optimized Restormer with a sparse Transformer and mixture of experts feature compensator (MEFC), leading to superior performance in de-raining tasks~\cite{drsformer}. However, Restormer’s Transformer generates keys and values from the entire feature map, resulting in high computational costs and longer inference times. Although DRS\red{f}ormer improves the Transformer, it has a large number of parameters and is specifically designed for de-raining scenarios.

In this paper, we propose a novel Transformer-based model that integrates the linear SRA mechanism from PVTv2~\cite{pvtv2} with CNN-based channel-wise attention, all within a gated UNet backbone. \modify{Our objective is to improve the removal of weather-related distortions from images by refining a Transformer-based approach, while simultaneously achieving a lightweight design that balances performance in terms of parameter size, computational overhead, memory usage, and inference time.} The linear SRA mechanism optimizes the Transformer by compressing the feature map, thereby reducing the size of the key and value generated during attention computation~\cite{attention}. The CNN-based channel-wise attention allows the Transformer to accommodate images of various sizes and better capture local features. Additionally, by incorporating a gating mechanism during downsampling, the model can selectively retain necessary information, minimizing the interference of extraneous data on model learning. This combination enables our model to accept images of various sizes and produce output images of the same size, demonstrating strong generalization capabilities. Experimental results indicate that our model performs impressively on raindrop, snow, and rain-fog datasets, proving its effectiveness in single and multiple weather removal tasks. The main contributions of our work can be summarized as follows.

\begin{enumerate}
\itemsep=0pt
  \item \red{We apply a single encoder-decoder architecture to remove various weather-related factors, including single- and multi-weather scenarios. This simplifies the model structure, reduces complexity, and significantly improves the generalisation ability and robustness of the model under different weather conditions.}
  \item \rsix{We employ PVTv2's linear SRA mechanism to boost inference speed and reduce resource consumption, ensuring efficient performance with consistent results.}
  \item \rsix{We integrate gating mechanisms into the downsampling process and feed-forward module in the Transformer to selectively preserve valuable information, enhancing feature extraction and ensuring accurate weather removal in complex scenes.}
  \item \rsix{We combine CNN-based channel-wise attention with the UNet architecture, enabling the model to process varying input image sizes while matching output dimensions to the input. This synergy significantly enhances local feature capture, enabling finer image detail restoration and superior de-weathering results.}
\end{enumerate}

\begin{figure*}[ht]
  \centering
  \includegraphics[width=\textwidth]{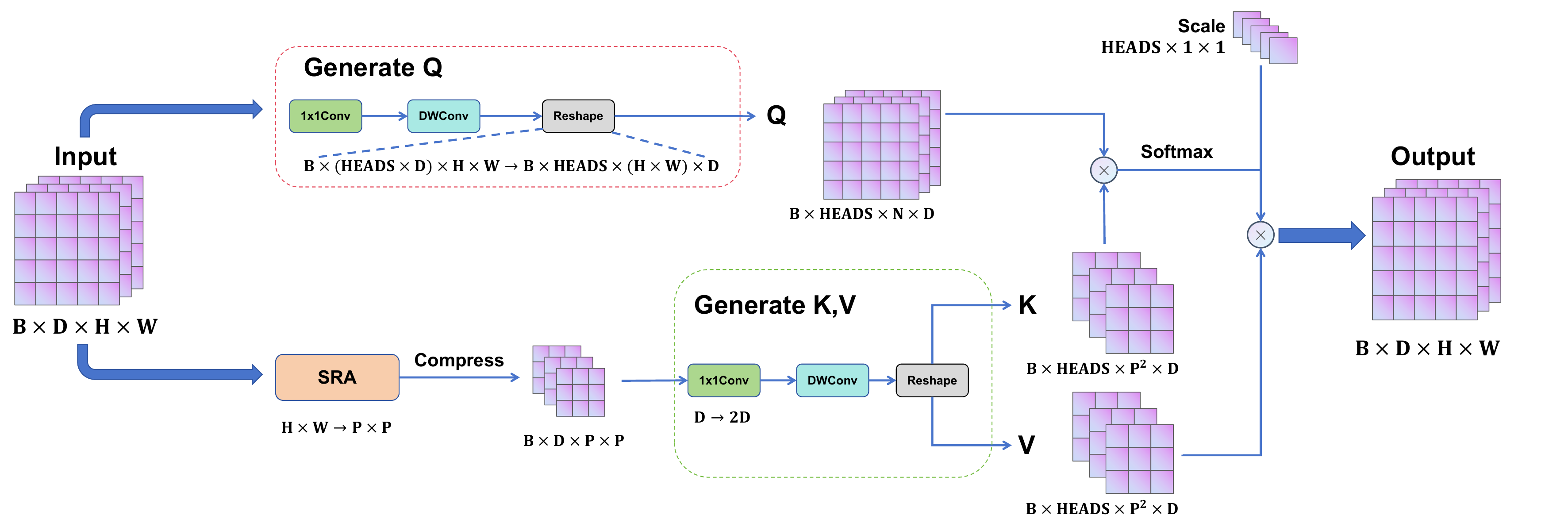}
  \caption{\red{\revise{The MSA architecture.} Within the MSA, the input matrix passes through a \(1 \times 1\) convolution layer and a \revision{depthwise separable convolution} to obtain the query. \revise{Next, adjust the dimensions of the query to adapt the computation of multi-head attention ("HEADS" represents the number of attention heads).} For the generation of the key and value, the input matrix is first compressed by linear SRA, reducing the dimensions from \(B \times D \times H \times W\) to \(B \times D \times P \times P\). It then passes through a \(1 \times 1\) convolution layer and a \revision{depthwise separable convolution} and is finally split along the channel dimension to obtain the key and value, which are adjusted to have the same dimensions as the query. The query, key, and value are then used to compute attention scores, which are scaled by a learnable parameter.}}
  \label{fig:pvtattn}
\end{figure*}

\section{Related Work}
In recent years, significant advancements have been made in developing effective techniques to counteract the negative impact of adverse weather conditions such as rain, fog, and snow on image quality. The field of weather restoration has evolved to include two main approaches: task-specific methods for single-weather removal and comprehensive frameworks for multi-weather removal.

\textbf{Single-weather Removal.} 
Task-specific methods involve training models independently for each type of weather-induced degradation, ensuring specialized and precise restoration for individual conditions. 
Early methods were primarily based on traditional image processing techniques. 
For instance, He \textit{et al.}~\cite{dark_channel_prior} introduced the Dark Channel Prior method to dehaze images by estimating atmospheric light and transmission maps. 
\revise{With the advent of deep learning, CNN-based methodologies have become increasingly prevalent in this domain.
Liu \textit{et al.}~\cite{durn} proposed DuRN, which employs paired and staggered skip connections in ResNet~\cite{resnet} to remove rain-streak and raindrops.}
\red{
Cui \textit{et al.}~\cite{focalnet} proposed FocalNet, which optimizes self-attention by aggregating contexts focally around and adaptively modulating the query with the aggregated context.
Afterwards, Cui \textit{et al.}~\cite{ConvIR} introduced \revise{multi-shape content-aware} attention modules that enhance informative high-frequency components in ConvIR.
More recently, Transformer-based models have been introduced in weather removal tasks.
Lin \textit{et al.}~\cite{LMQFormer} employs Laplace-VQVAE in LMQFormer to generate a coarse mask as prior knowledge, thereby improving the fine-grained details of desnowing.
Moreover, Chen \textit{et al.}~\cite{drsformer} designed DRSformer with the learnable \revise{Top-K} selection and a mixed-scale \rfive{feed-forward network (FN)} to improve feature aggregation.}

\textbf{Multi-weather Removal.}
All-in-one methods aim to develop a unified model capable of addressing multiple adverse weather conditions, simultaneously, offering a versatile and comprehensive solution. These advancements not only enhance image clarity and reliability under challenging conditions, but also pave the way for more robust visual systems. Li
\textit{et al.}~\cite{allinone} introduced the All-in-One approach, the first model for multi-weather removal. This method leverages a multi-task learning strategy to share representations across various weather conditions.
\red{To construct a unified framework, Zamir \textit{et al.}~\cite{restormer} proposed Restormer, which integrates ViT into a UNet architecture with transposed self-attention, achieved notable performance on several tasks including rain removal.} Jose Valanarasu \textit{et al.}~\cite{transweather} designed TransWeather, a transformer-based architecture with learnable weather-type queries, incorporating sub-patches to remove small-scale weather degradations.
Recently, Özdenizci and Legenstein~\cite{wd} introduced WeatherDiffusion, which innovatively leverages a guided denoising diffusion process~\cite{ddpm} with smoothed noise estimates across overlapping patches to restore images.

\textbf{Vision Transformer.}
The Transformer architecture, originally designed for natural language processing tasks, was first introduced to computer vision by Dosovitskiy \textit{et al.}~\cite{vit}, which demonstrated a superior ability to capture long-range dependencies and contextual information more effectively than CNNs. 
Since its introduction, ViT has been widely adopted for various computer vision tasks, including image and video recognition~\cite{timesformer}, segmentation~\cite{sa}, and object detection~\cite{detr}. 
However, \revise{since} ViT relies on self-attention~\cite{attention}, its computational complexity can increase with the number of image patches. To address this, recent research has focused on enhancing the ViT architecture. Liu \textit{et al.}~\cite{swin} developed the Swin Transformer, which employs a hierarchical structure and shifted windows to improve computational efficiency and scalability. Wang \textit{et al.}~\cite{pvt} proposed PVT and PVTv2, further advancing the ViT architecture by incorporating a pyramid structure and SRA, reducing time complexity to linear.

\red{
\textbf{Comparison with \revision{Existing Works}.}
Currently, most weather removal models built on the Transformer architecture, such as Restormer and TransWeather, generate query, key, and value from the entire feature map within the attention mechanism. This method introduces excessive information, complicating the model's ability to learn features suited for various scenarios. \rfive{Although DRSformer employs a sparsification technique, its FN design yields a large number of parameters and excessive information overload~\cite{drsformer}.} Our model addresses these challenges by employing a gating mechanism alongside a linear SRA mechanism within convolutional attention. This approach focuses on extracting only the essential features, allowing the model to adapt effectively to diverse weather conditions while minimizing parameters and resource overhead.
\rfive{Additionally, while WeatherDiffusion has demonstrated strong performance across various weather conditions, our model offers a notable advantage. Unlike WeatherDiffusion, which processes images in segments due to its design~\cite{wd}, our convolution-based attention mechanism enables direct processing of entire images. This reduces inference time significantly. Moreover, diffusion models rely on iterative denoising steps to generate data, meaning that inferring a single image often requires multiple iterations, leading to considerable computational overhead~\cite{bs-ldm}. In comparison, our model, like other Transformer-based approaches, completes inference in a single iteration, making it far more computationally efficient.}
}

\section{Network Architecture}
Due to Restormer's~\cite{restormer} use of the entire feature map for attention score computation, it exhibited longer inference times and contained a relatively large number of parameters. To address this issue, we employed a modified PVTv2~\cite{pvtv2}, called a multi-scale pyramid vision transformer (MS-PVT), to optimize inference time and reduce parameter count. Given MS-PVT’s suitability for feature pyramids, a UNet-type \cite{unet} structure remains appropriate.

The architecture of WeatherRemover is shown in Fig.~\ref{fig:backbone}. Initially, a convolutional layer processes a degraded image $\textbf{I} \in \mathbb{R}^{3 \times H \times W}$ to extract low-level feature embeddings $\textbf{F}_0 \in \mathbb{R}^{D \times H \times W}$, where $D$ represents the feature dimension and $H \times W$ denotes the spatial dimensions. \red{Subsequently, $\textbf{F}_0$ is fed into a three-stage symmetric encoder-decoder backbone. Each stage consists of a sequence of MS-PVT Blocks and sampling layers to extract multi-scale features.} 

The left part functions as \revise{an} encoder, responsible for downsampling the input features. At each downsampling stage (D-Stage), a gating mechanism is employed to select necessary information and eliminate the influence of redundant data on the model's learning process. \rsix{In particular, one branch learns a matrix of scaling weights, which interact element-wise with weights from the other branch. This interaction selectively adjusts the information, reducing redundancy and improving downsampling efficiency.}
The input features \revise{$\textbf{F}_k$} are divided into two parallel paths: the main path $P_m$ and the bypass path $P_b$. The bypass path consists of a single MS-PVT block, while the main path comprises a sequence of MS-PVT blocks. \revise{The number of blocks $n$ varies at different stages}, enabling the model to extract latent features more effectively. The encoding process at each stage can be summarized as follows:
\revise{
\begin{equation}
  \label{pvtblock}
  \begin{aligned}
    P_m &= \mathrm{MSPVT}_n(...(\mathrm{MSPVT}_2(\mathrm{MSPVT}_1(\textbf{F}_k)))), \\
    P_b &= \mathrm{MSPVT} (\textbf{F}_k),~\textbf{F}_{k+1} = \textbf{Down}(\phi(P_m) \odot P_b),
  \end{aligned}
\end{equation}
}
where $\odot$ denotes element-wise multiplication, $\phi$ represents the non-linearity activation~\cite{gelu}, and \revise{\textbf{Down}} is the downsampling operation.

The decoder takes low-level latent features $\textbf{F}_l \in \mathbb{R}^{8D \times \frac{H}{8} \times \frac{W}{8}}$ as input for high-level image restoration. At each stage, features from both the encoder and decoder are aggregated across channels in the fusion layer via skip connections~\cite{unet}. To downsample and upsample features, WeatherRemover employs pixel-unshuffle and pixel-shuffle operations, respectively~\cite{pixel_shuffle}. 

After the decoder stage, the deep features $\textbf{F}_n$ are further enriched at the Refinement stage to generate a residual image $\hat{\textbf{R}} \in \mathbb{R}^{3 \times H \times W}$, which is added to the original input $\textbf{I}$ to obtain the final restored image $\hat{\textbf{I}}=\textbf{I}+\hat{\textbf{R}}$. This approach helps preserve fine-grained structural and textural details in the restored output.

\begin{figure}[ht]
  \centering
  \includegraphics[scale=0.4]{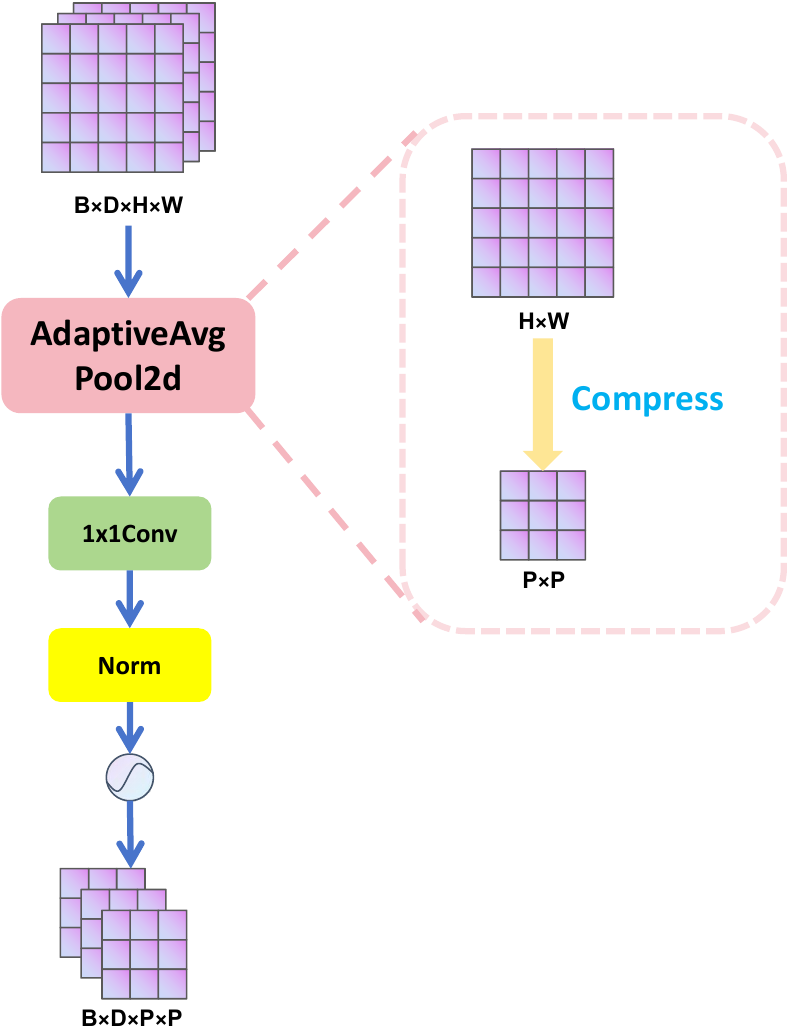}
  \caption{\red{\revise{The linear SRA flowchart for compressing feature maps.} The primary component is an adaptive mean pooling layer. \revise{It compresses the tensor from dimensions \(H \times W\) to \(P \times P\), where \(P\) represents the size of the feature map output from the adaptive average pooling layer.} The input tensor, with an initial dimension of \(B \times D \times H \times W\), is transformed to \(B \times D \times P \times P\) after passing through the adaptive mean pooling layer. Subsequently, it is processed by a \(1 \times 1\) convolutional layer and normalized. Finally, the tensor is activated by an activation function.
}}
  \label{fig:sra}
  \vspace{-0.5cm}
\end{figure}

\subsection{Multi-Scale Attention Structure}
Since the backbone needs to handle varying input image sizes, the original PVT and PVTv2 cannot directly perform the patch operation. To address this, we propose the multi-scale attention (MSA) shown in Fig.~\ref{fig:pvtattn}, which replaces all the linear layers in the original PVT Attention with $1 \times 1$ convolution ($W^p$) to aggregate pixel-wise cross-channel context and $3 \times 3$ depthwise separable convolution ($W^d$) to encode spatial information, yielding the transformation from input \textbf{X} to query (\textbf{Q}) as:
\rsix{
\begin{equation}
  \label{q}
  \revision{\widehat{\textbf{Q}}=W^d_qW^p_q\textbf{X}}.
\end{equation}
}
In PVTv2 \cite{pvtv2}, the core module is the linear SRA \red{(shown in Fig.~\ref{fig:sra})}. \red{Linear SRA uses the adaptive average pooling layer to compress the feature map $F \in \mathbb{R}^{D \times H \times W}$ into $F' \in \mathbb{R}^{D \times P \times P}$, with $P$ representing the pooling size of linear SRA~\cite{pvtv2}.} This is followed by a $1 \times 1$ convolution layer for linear transformation, then normalization and activation. The function of linear SRA is described as:
\rsix{
\begin{equation}
  \label{sra}
  \revision{\mathrm{SRA}(\textbf{X}) = \phi(\lambda(W^p_c \textbf{X}))},
\end{equation}
}where $\lambda$ indicates layer normalization and $\phi$ denotes non-linearity activation. The key (\textbf{K}) and value (\textbf{V}) are generated similarly to the query but with linear SRA applied as:
\rsix{
\begin{equation}
  \label{kv}
    \revision{\widehat{\textbf{K}}= W^d_kW^p_k\mathrm{SRA}(\textbf{X})}, 
    \revision{\widehat{\textbf{V}}= W^d_vW^p_v\mathrm{SRA}(\textbf{X})}. \\
\end{equation}
}
Overall, the MSA can be described as:
\rsix{
\begin{equation}
    \label{msa}
    \revision{\mathrm{MSA}(\textbf{X}) = W^p_{out}(\sigma(\widehat{\textbf{Q}}\widehat{\textbf{K}}^{\mathsf{T}} \textbf{A})\widehat{\textbf{V}})},
\end{equation}
}where $\sigma$ denotes the softmax operation and \textbf{A} is a learnable temperature control coefficient, which helps prevent the vanishing gradient problem~\cite {attention}. When disabled, it defaults to $\sfrac{1}{\sqrt{D}}$.

\subsection{Gating Feed-Forward Network}
\begin{figure}[ht]
    \centering
    \includegraphics[scale=0.55]{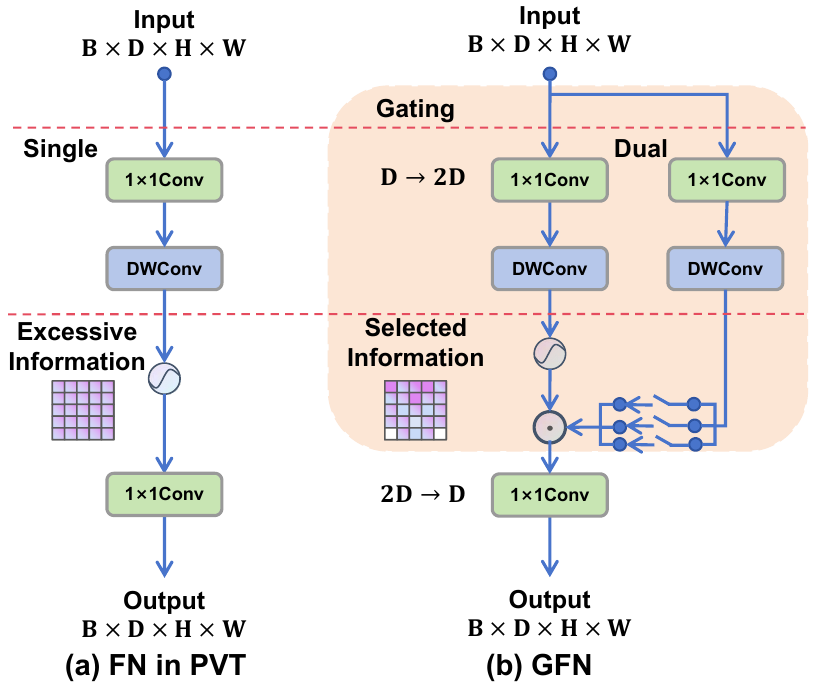}
    \caption{\red{\revise{Comparison between PVTv2 and GFN structures.}} \textbf{(a)} depicts the structure of PVTv2, and \textbf{(b)} illustrates our proposed structure GFN. \rfinal{PVTv2’s serial FN structure was modified with a gating mechanism: the original single-path 1x1 and depthwise separable convolution were split into dual paths, one activated, then fused via element-wise multiplication.}}
    \label{fig:mlp}
\end{figure}
Fig.~\ref{fig:mlp} compares the FN of PVTv2 with our structure. The original FN in PVTv2 utilized a single pathway (Fig.~\ref{fig:mlp}a), which inadvertently allowed irrelevant information to affect the model's learning process.

In contrast, our enhanced network \modify{gating feed-forward network (GFN)} (Fig.~\ref{fig:mlp}b) incorporates a gating mechanism to selectively filter the information transmitted, thereby reducing disruptions caused by extraneous data. The gating mechanism comprises two identical branches. One branch is activated using an activation function, and the outputs of the two branches are then multiplied element-wise. For a given input \textbf{X}, our GFN is formulated as follows:
\begin{equation}
  \label{ffn}
  \begin{aligned}
    \mathrm{GFN}(\textbf{X}) &= W^p_{out} \mathrm{Gating} (W^p_{in}\textbf{X}), \\
    \mathrm{Gating}(W^p_{in}\textbf{X}) &= \phi(W^d_1W^p_1W^p_{in}\textbf{X}) \odot W^d_2W^p_2W^p_{in}\textbf{X},
  \end{aligned}
\end{equation} where $\phi$ represents the non-linearity activation.

\subsection{MS-PVT Block Structure}
\begin{figure}[ht]
  \centering
  \includegraphics[scale=0.35]{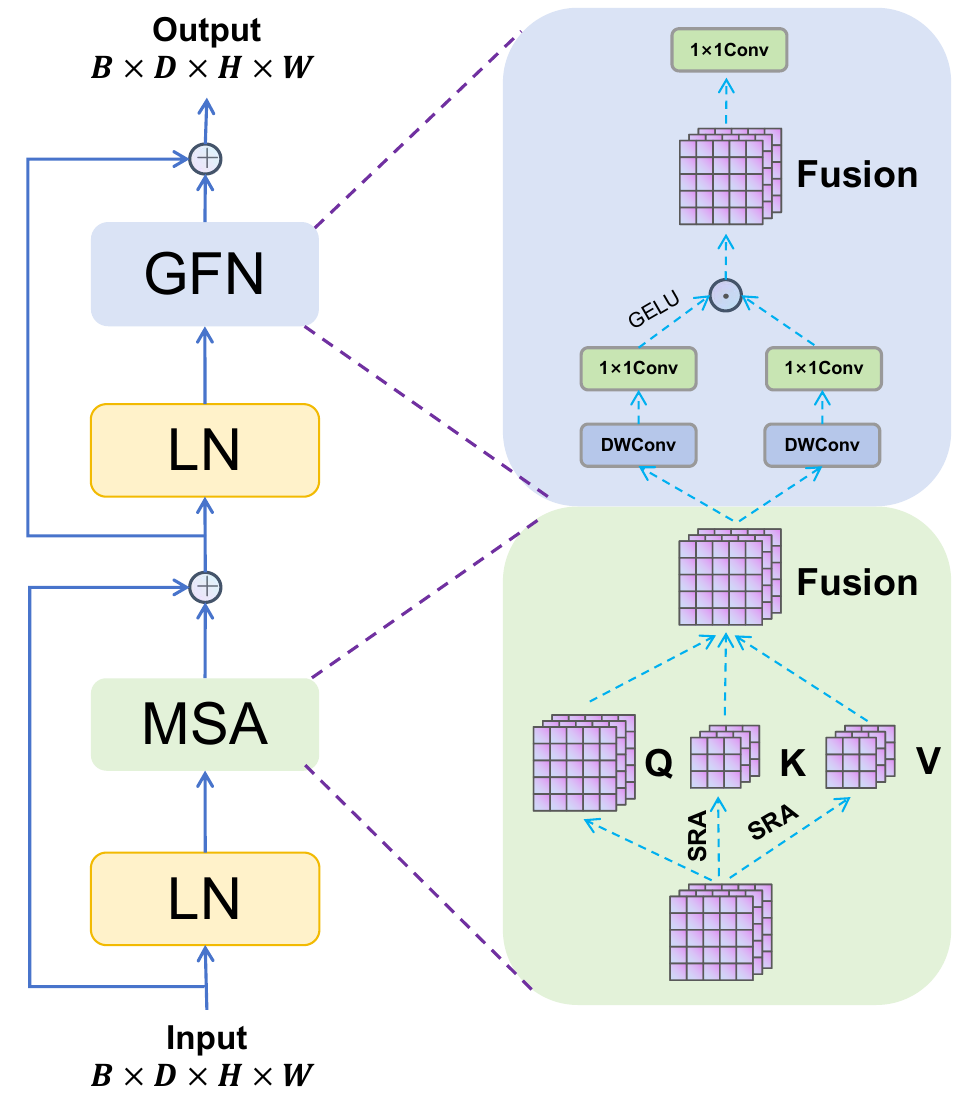}
  \caption{\red{\revise{Overall structure of MS-PVT.} MS-PVT is composed of MSA and GFN in series. The outputs of both MSA and GFN are normalized and added to the input. \revision{On the right side of both GFN and MSA, there are overall descriptions of the internal data flow.} For an input with dimensions \(B \times D \times H \times W\), the output also has dimensions \(B \times D \times H \times W\).}}
  \label{fig:block}
\end{figure}

The fundamental structure of MS-PVT is divided into two main components, as shown in Fig.~\ref{fig:block}. These components are the MS-PVT module and the GFN. After processing through each component, an addition operation is performed with the respective input. It is important to note that the MS-PVT module maintains the dimensionality of both its inputs and outputs.

Initially, the input \textbf{X} undergoes layer normalization, followed by the MS-Attention mechanism. The result is then added to the original input, denoted as $\widehat{\textbf{X}}$. Subsequently, it passes through a GFN, where the outputs are added to the inputs before entering the final layer normalization. This process can be described as:
\begin{equation}
  \label{block}
  \begin{aligned}
    \widehat{\textbf{X}} &= \textbf{X}+\mathrm{MSA}(\lambda(\textbf{X})), \\
    \mathrm{MSPVT}(\textbf{X}) &= \widehat{\textbf{X}}+\mathrm{GFN}(\lambda(\widehat{\textbf{X}})),
  \end{aligned}
\end{equation} where $\lambda$ indicates layer normalization.

\red{In the MS-PVT framework, the linear SRA within the MSA module compresses the key and value components to preserve essential features while minimizing computational load, maintaining effectiveness. However, because the query remains uncompressed, it can retain redundant information. To address this, the subsequent GFN module employs a gating mechanism to learn scaling weights. These weights are integrated with the backbone model through element-wise multiplication, effectively enhancing relevant information and suppressing irrelevant details. This approach reduces the impact of redundant data on the model's learning process.}

\modify{To address the challenges posed by extreme weather conditions, such as heavy snow and dense fog, which obscure critical details in images, the linear SRA mechanism in MS-PVT compresses multi-scale feature maps, preserving only essential structural information and mitigating weather-induced interference. Simultaneously, the GFN leverages a gating mechanism to selectively retrieve and enhance useful image features while suppressing irrelevant elements, enabling progressive removal of dense fog and heavy snow, and significantly improving image restoration quality.}

\subsection{Time Complexity Analysis}
\rsix{\red{In this section, we evaluate the computational complexity of each component in our model. The $1 \times 1$ convolution, represented as \(W^p\), exhibits a computational complexity of \(\mathcal{O}(D^2 \cdot K^2 \cdot \textbf{X})\). In contrast, the $3 \times 3$ depthwise separable convolution, denoted as \(W^d\), has a complexity of \(\mathcal{O}(D \cdot \textbf{X} \cdot (D+K^2))\). Given that the dimension \(D\) and kernel size \(K\) are predetermined hyperparameters, these complexities simplify to \(\mathcal{O}(\textbf{X})\). Additionally, the computation of the attention score has a time complexity of \(\mathcal{O}(\textbf{X}^2)\), resulting in the MSA mechanism also having a complexity of \(\mathcal{O}(\textbf{X}^2)\). Furthermore, as demonstrated in Eq. (\ref{ffn}), the operations involving \(W^p\) and \(W^d\) within the GFN module each have a complexity of \(\mathcal{O}(\textbf{X})\). Consequently, the overall time complexity of the MS-PVT block is \(\mathcal{O}(\textbf{X}^2)\).}}

\subsection{Loss Function}
Commonly used loss functions in deep learning include the $\mathcal{L}_1$ and $\mathcal{L}_2$ losses. The $\mathcal{L}_1$ loss, also known as the mean absolute error (MAE), is less sensitive to outliers because it measures the error in absolute terms, although it is non-differentiable at discontinuities. Conversely, the $\mathcal{L}_2$ loss function, also known as the mean squared error (MSE), has a relatively stable derivative, which helps models converge by maintaining a consistent gradient. However, $\mathcal{L}_2$ loss is prone to falling into local optima and is more influenced by outliers.

To combine the strengths of both the $\mathcal{L}_1$ and $\mathcal{L}_2$ loss functions, we employed the Pseudo-Huber metric as proposed by Song and Dhariwal~\cite{cm_improved} in their work on the consistency model. The Pseudo-Huber loss is defined as:
\begin{equation}
  \mathcal{L}_{pHuber}(\textbf{Y},\hat{\textbf{I}}) = \sqrt{\| \textbf{Y},\hat{\textbf{I}} \|^2_2 + c^2} - c,
\end{equation}
where \textbf{Y} represents the ground truth and $c>0$ is an adjustable constant.

\section{Experiments}
\subsection{Datasets}
We trained and evaluated the WeatherRemover on three standard benchmark datasets for adverse weather image restoration, specifically targeting the removal of snow, rain, and raindrops on camera sensors. Additionally, we used a combined dataset for an All-in-one Weather Removal assessment.

\textbf{Snow100K~\cite{desnownet-snow100k}.} 
The Snow100K dataset consists of synthetic images showing snow-covered scenes with varying snowfall intensities. The dataset is divided into a training set of 50,000 images and a test set that is further segmented into three subsets: Snow100K-S (Small), Snow100K-M (Medium), and Snow100K-L (Large). These subsets contain 16,611, 16,588, and 16,801 images, respectively, with snow intensity increasing from Snow100K-S to Snow100K-L. For experimental evaluations, we used the entire Snow100K-L test set to assess our models. \red{The dataset comprises 1,329 realistic images depicting snowy conditions. These images are essential for assessing the model’s effectiveness in dealing with real-world environments.}

\textbf{RainDrop~\cite{attentivegan-raindrop}.} 
The RainDrop dataset contains images that exhibit degradation due to raindrop artifacts on camera sensors, which obscure the fine details of background objects. The dataset is composed of a training set with 861 images and a test set that is subdivided into RainDrop-A and RainDrop-B. The former comprises 58 pairs of images, while the latter includes 249 pairs. Notably, RainDrop-A represents a subset of RainDrop-B, featuring carefully aligned image pairs that have been extracted from the latter.

\textbf{Outdoor-Rain~\cite{hrgan}.} 
The Outdoor-Rain dataset simulates mixed rain and fog scenes and contains 600 sets of images. Each set comprises 15 degraded images with varying levels of rain and haze and one clean image as ground truth. We used the last 50 sets (750 images) as the test set and the remaining sets for training.

\textbf{All-Weather~\cite{transweather}.} 
The All-Weather dataset is a combination of the training sets from the aforementioned datasets, comprising 9,000 images sampled from Snow100K, 1,069 images from Raindrop, and 9,000 images from Outdoor-Rain.

\red{\textbf{DAWN~\cite{dawn}.}}
\red{The Detection in Adverse Weather Nature (DAWN) dataset comprises 1,000 real-world traffic images captured in challenging weather conditions, including fog, snow, rain, and sandstorms. We utilized this dataset to assess our model's restoration performance specifically in real-world rain and fog scenarios.}

\subsection{Implementation Details}
We classify the tasks into two categories: single-weather tasks and multi-weather mixing tasks. Single-weather tasks involve training and evaluating models on a single dataset. For mixing tasks, we trained on the All-Weather dataset~\cite{transweather} and evaluated on selected corresponding test sets. In the three stages of downsampling, we utilized 4, 6, and 6 MS-PVT blocks, respectively, with only one MS-PVT block in the gated branch at each stage. The number of attention heads for these stages was set to 4, 8, and 12, respectively. At each stage, the channel count was doubled while the height and width were halved. For the fourth stage, we employed 8 MS-PVT blocks, each with 16 heads, and did not use the gating mechanism. In the three stages of upsampling, the configuration mirrored that of the downsampling stages, but the channel count at each stage was halved, and the height and width were doubled. During the Refinement stage, we used 4 MS-PVT blocks, each with 4 heads. The linear SRA size for all MS-PVT modules was set to 7, with a learnable temperature control coefficient enabled for each module. In the loss function, we set \(c\) to 0.03.

We conducted our experiments using PyTorch version 2.0.0, training and evaluating the model on a single RTX 3090 GPU with the Adam optimizer~\cite{adam}. \revise{To accommodate different image sizes, we began each task with a batch size of 8 and an image size of 128$\times$128, starting with an initial learning rate of 0.0003}. Throughout 800,000 iterations, we gradually reduced the batch size to 1, increased the image size to \revise{320$\times$320}, and lowered the learning rate to 0.0000001.

\subsection{Weather Removal Results Analysis}

\begin{figure*}[ht]
  \centering
  \includegraphics[scale=0.4]{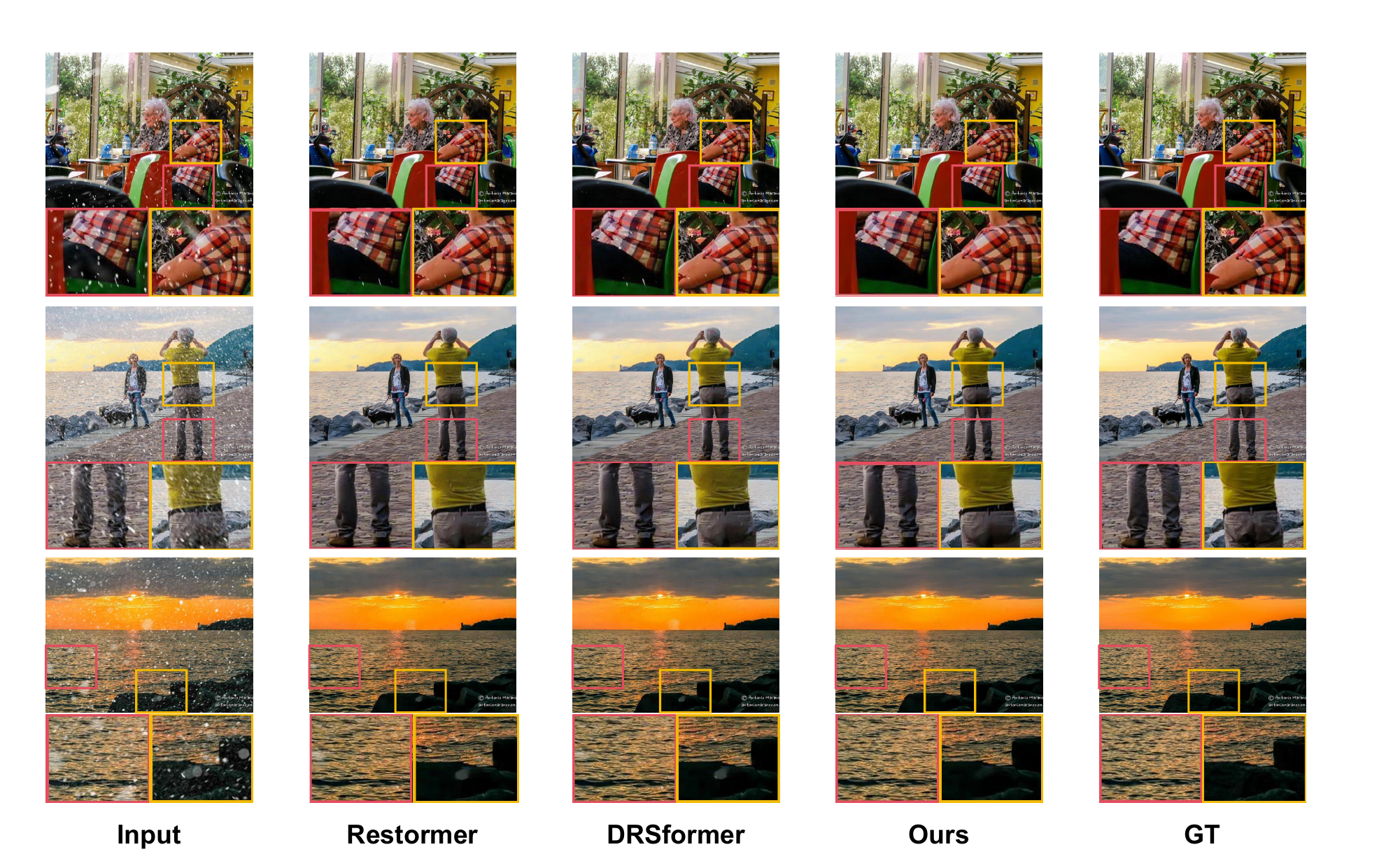}
  \caption{\red{\revise{Desnowing visual comparisons.}} From left to right, the sequence of images includes the input images, Restormer~\cite{restormer}, DRS\red{f}ormer~\cite{drsformer}, our model, and the ground truth images. Each image is accompanied by magnified details beneath it for closer examination. \rsix{Images are selected from the Snow100K-L~\cite{desnownet-snow100k} test set, focusing on scenes where the background color closely resembles the snowflake color.}}
  \label{fig:snow100k}
\end{figure*}

\textbf{Desonwing results.} As shown in Table~\ref{bk-s100k}, we evaluated models trained on the Snow100K~\cite{desnownet-snow100k} and AllWeather~\cite{transweather} datasets using the Snow100K-L. We compared models in recent years in terms of peak signal-to-noise ratio (PSNR)~\cite{psnr}, structural similarity (SSIM)~\cite{ssim}, inference time, and parameter count. The results indicate that our model achieved the highest performance in terms of PSNR under both single and multiple weather conditions.

\begin{figure*}[h]
  \centering
  \includegraphics[scale=0.4]{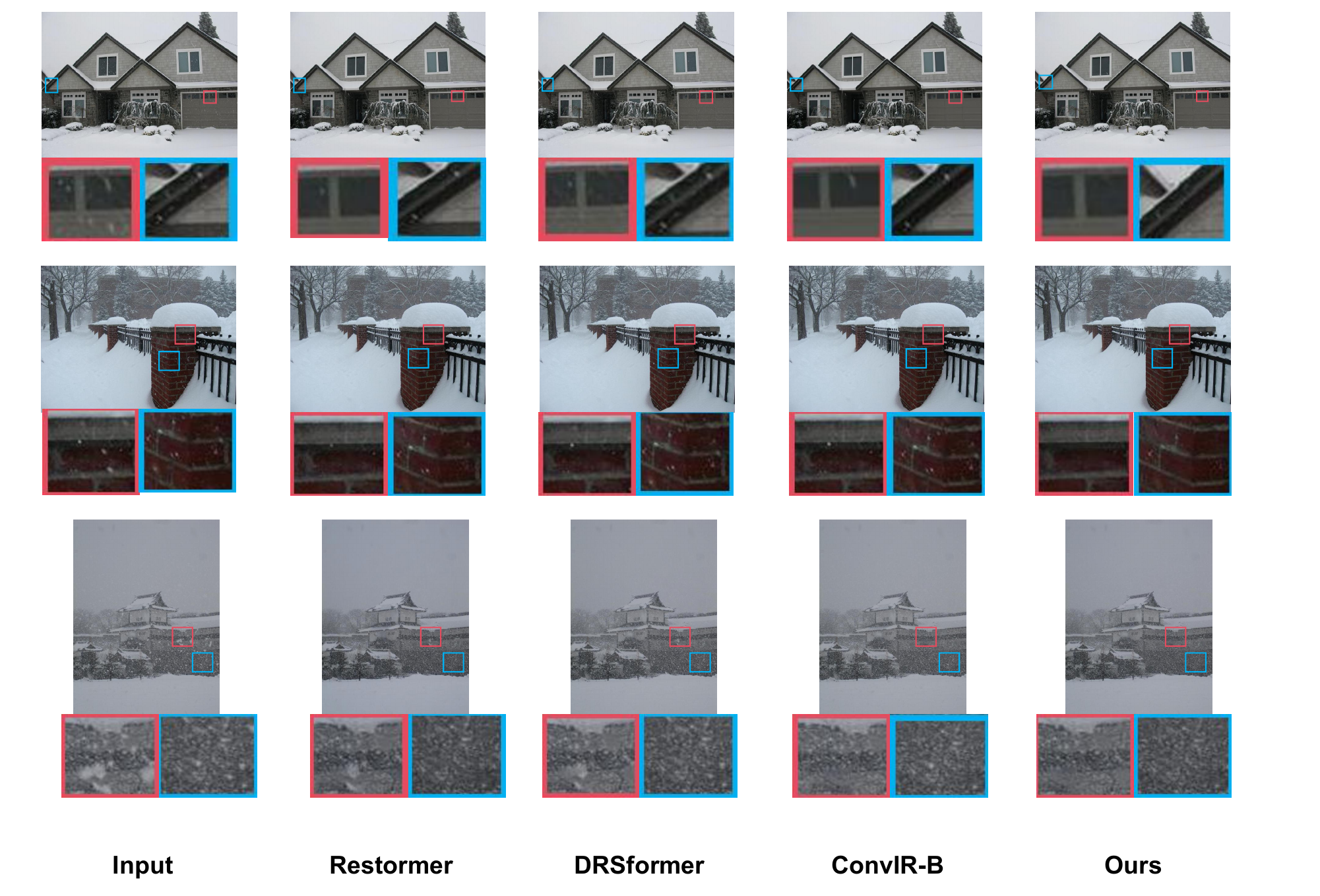}
  \caption{\red{\revise{Real-scene desnowing visual comparisons.} The figures from left to right are: Input images, Restormer~\cite{restormer}, DRSformer~\cite{drsformer}, ConvIR-B~\cite{ConvIR}, and our model. The zoomed-in details are shown below each image. \rsix{Images were selected from the DAWN~\cite{dawn} dataset. The first two rows demonstrate the models' ability to restore fine details in real-world scenarios, while the third row highlights the performance of the models in removing heavy snow in low-resolution settings, aiming to reveal our limitations under such conditions.}}}
  \label{fig:snow_real}
\end{figure*}

Under single weather conditions, although our model’s SSIM is 0.0037 dB lower than that of ConvIR-B~\cite{ConvIR}, we surpass it by 1.7 dB in PSNR. The SSIM metric is particularly attuned to the structural integrity of images, whereas the PSNR metric is more responsive to variations in brightness and contrast. This suggests that while our model exhibits minor shortcomings in reconstructing snowy image structures for single-weather snow removal tasks, it excels in enhancing contrast and brightness. Fig.~\ref{fig:snow100k} illustrates the visual performance of various models in removing snow from images under single-weather conditions. A closer examination of the magnified details reveals that both Restormer~\cite{restormer} and DRS\red{f}ormer~\cite{drsformer} struggle to preserve fine details. In contrast, our model demonstrates superior ability in managing minute snow traces and snowflakes. Notably, our model effectively eliminates snow traces that blend in color with the background and accurately reconstructs the underlying scene using contextual information from the surroundings. When compared to the ground truth, the images restored by our model exhibit a high degree of similarity to the original, unaltered images.

\red{Additionally, we assessed our model's performance in removing snow from images captured in real snowy conditions. The visual comparisons are presented in Fig.~\ref{fig:snow_real}. The close-up details in the first and second rows of images reveal that our model excels at eliminating small snow traces, indicating a strong ability to capture fine details. \modify{However, as demonstrated in the third-row image, our model struggles to effectively restore images with dense snow under low-resolution conditions. The magnified details show that while large snowflakes are partially removed, dense traces of snow persist. This highlights deficiencies in the model’s ability to integrate contextual information, particularly in real-world dense snowfall scenarios. Future work should prioritize enhancing its capacity for longer-range contextual integration.}}

Although LMQFormer~\cite{LMQFormer}, ConvIR~\cite{ConvIR}, and FocalNet~\cite{focalnet} are very lightweight, they are only suitable for single-weather scenarios. Furthermore, LMQFormer is specifically designed for snow removal tasks. Consequently, in our multi-weather removal experiments, we compared models suited for multiple weather conditions, namely WeatherDiff$_{64}$, WeatherDiff$_{128}$, Restormer, and TransWeather. The results show that while our model's inference time is about 18\% slower than TransWeather, it improves PSNR by 10\% and achieves SSIM comparable to Restormer. These findings demonstrate that our model performs well in mixed-weather scenarios, offering both effectiveness and efficiency.

\begin{table}[h]
  \caption{Desnowing results on Snow100K-L.}
  \centering
    \resizebox{\linewidth}{!}{
  \begin{tabular}{llllll}
    \hline
    \textbf{\hspace{0.15cm}Task} & Model & PSNR$\uparrow$ & SSIM$\uparrow$ & Time/s & Param \\
    \hline
    \multirow{10}{*}{\centering \shortstack{Snow\\Removal}}
    & SnowDiff$_{64}$~\cite{wd}           & 30.43 & 0.9145 &\hspace{0.1cm} \textgreater10  & 82.9M \\
    & SnowDiff$_{128}$~\cite{wd}          & 30.28 & 0.9000 &\hspace{0.1cm} \textgreater10  & 85.6M \\
    & Restormer~\cite{restormer}          & 31.48 & 0.9286 &\hspace{0.1cm} 0.15 & 26.5M \\
    & DRS\red{f}ormer~\cite{drsformer}          & 31.06 & 0.9100 &\hspace{0.1cm} 0.21 & 33.6M \\
    & LMQFormer~\cite{LMQFormer}          & 31.24 & 0.9100 &\hspace{0.1cm} 0.02 & 2.18M \\
    & ConvIR-S~\cite{ConvIR}              & 30.49 & 0.9290 &\hspace{0.1cm} 0.02 & 5.53M \\
    & ConvIR-B~\cite{ConvIR}              & 30.56 & \redbf{0.9305} &\hspace{0.1cm} 0.02 & 8.63M \\
    & FocalNet~\cite{focalnet}            & 30.15 & 0.9270 &\hspace{0.1cm} 0.01 & 3.74M \\
    & \textbf{WeatherRemover (Ours)}       & \redbf{32.26} & 0.9268 &\hspace{0.1cm} 0.11 & 24.3M \\
    \hline
    \multirow{5}{*}{\centering \shortstack{Multi\\Weather}}
    & TransWeather~\cite{transweather}  & 27.95 & 0.9037 &\hspace{0.1cm} 0.09 & 38.1M \\
    & WeatherDiff$_{64}$~\cite{wd}       & 30.09 & 0.9041 &\hspace{0.1cm} \textgreater10  & 82.9M \\
    & WeatherDiff$_{128}$~\cite{wd}     & 29.58 & 0.8941 &\hspace{0.1cm} \textgreater10  & 85.6M \\
    & Restomer~\cite{restormer}         & 30.71 & \redbf{0.9131} &\hspace{0.1cm} 0.15 & 26.5M \\
    & \textbf{WeatherRemover (Ours)}     & \redbf{30.87} & 0.9121 &\hspace{0.1cm} 0.11 & 24.3M \\
    \hline
  \end{tabular}
  }
    \label{bk-s100k}
    \begin{flushleft}
    Note: The experiment is divided into two parts: single-weather desnowing and multi-weather desnowing. The text marked in bold red indicates the best PSNR or SSIM values in each part.
  \end{flushleft}
  \vspace{-0.5cm}
\end{table}

\textbf{Raindrop removal results.} As shown in Table~\ref{tb-rd}, we evaluated models trained on the Raindrop~\cite{attentivegan-raindrop} and AllWeather~\cite{transweather} datasets using the Raindrop-A. The results indicate that our model achieves state-of-the-art performance under single-weather conditions. Moreover, our PSNR is the highest under multiple weather.

\begin{figure*}[htbp]
  \centering
  \includegraphics[scale=0.4]{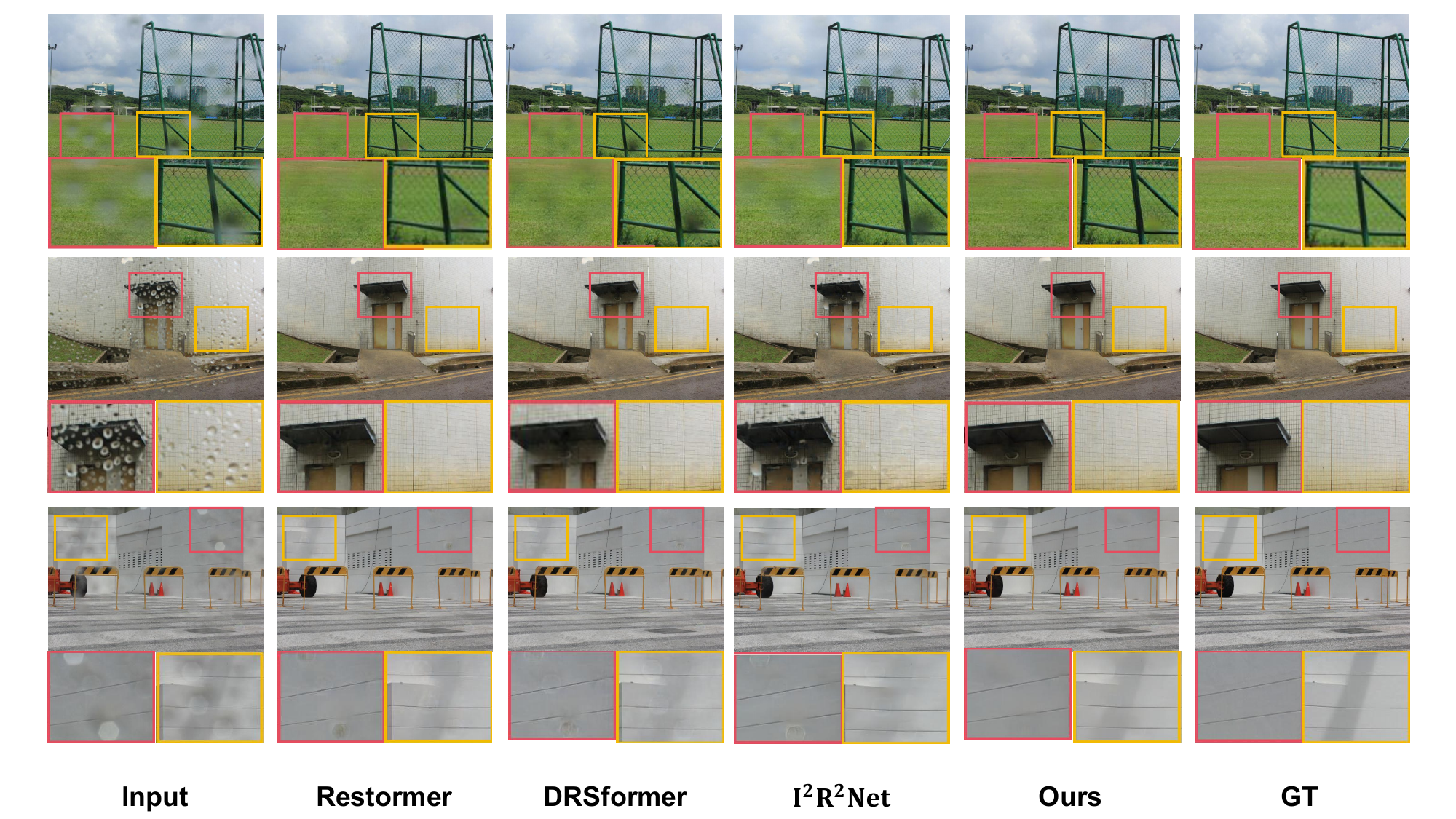}
  \caption{\red{\revise{Raindrop removal visual comparisons.}} From left to right, the images are: \red{Input images}, Restormer~\cite{restormer}, DRS\red{f}ormer~\cite{drsformer}, I$^2$R$^2$Net~\cite{i2r2net}, our model, and ground truth images. Each image includes magnified details below. \rsix{The images are from the RainDrop~\cite{attentivegan-raindrop} dataset's test set. The first row showcases restoration on backgrounds resembling rain droplet colors, the second highlights dense droplet removal, and the third contrasts models' impacts on texture and detail preservation.}}
  \label{fig:raindrop}
\end{figure*}

In single-weather scenarios, RainDropDiff, which is based on diffusion models, suffers from long inference times. Our model, with approximately 70\% fewer parameters, outperforms RainDropDiff in both PSNR and SSIM while also offering relatively shorter inference times. Although I$^2$R$^2$Net, which employs \red{long short-term memory} (LSTM) architecture, has the fewest parameters among these models, our method improves PSNR by 4.25 dB and SSIM by 0.0431. The LSTM-based architecture in I$^2$R$^2$Net introduces significant computational burden due to the gating mechanisms and the process of updating each LSTM's cell state and hidden state~\cite{lstm}, resulting in relatively long inference times. Compared to DuRN, which has the shortest inference time among the evaluated models, our model achieves a PSNR that is 1.75 dB higher. Fig.~\ref{fig:raindrop} offers a visual comparison of various models' performance on the RainDrop-A under single-weather conditions. Upon closer examination of the enlarged details, our model's rendition of the background aligns more closely with the ground truth than the results from other models. Furthermore, the other models fail to fully eliminate large raindrops, leaving behind noticeable shadows where the raindrops originally appeared. In stark contrast, our model effectively removes large raindrops, leaving a cleaner image.

Under multi-weather scenarios, our model's inference time is approximately 0.15 seconds longer than ~TransWeather~\cite{transweather}. However, it improves PSNR by about 11.8\% and has a smaller parameter count. Additionally, our model is more efficient compared to WeatherDiff$_{64}$, featuring both a smaller parameter count and a shorter inference time, while its SSIM is only marginally lower by 0.0005 dB, demonstrating its effectiveness. Based on the experimental results from both single-weather and multi-weather raindrop removal tasks, our model excels at reconstructing image structure and enhancing brightness and contrast.

\begin{table}[]
  \caption{Raindrop removal results on RainDrop-A.}
  \label{tb-rd}
    \resizebox{\linewidth}{!}{
  \centering
  \begin{tabular}{llllll}
    \hline
    \textbf{\hspace{0.15cm}Task} & Model & PSNR$\uparrow$ & SSIM$\uparrow$ & Time/s & Param \\
    \hline
    \multirow{11}{*}{\centering \shortstack{Raindrop\\Removal}}
    & AttentiveGAN~\cite{attentivegan-raindrop} & 31.59 & 0.9170 &\hspace{0.1cm} 3.21 & 6.2M  \\
    & DuRN~\cite{durn}                          & 31.24 & 0.9259 &\hspace{0.1cm} 0.01 & 10.2M \\
    & RaindropAttn~\cite{raindropattn}          & 31.44 & 0.9263 &\hspace{0.1cm} 0.19 & 7.3M   \\
    & IDT~\cite{idt}                            & 31.87 & 0.9313 &\hspace{0.1cm} 1.05 & 16.4M \\
    & RainDropDiff$_{64}$~\cite{wd}             & 32.29 & 0.9422 &\hspace{0.1cm} \textgreater10  & 82.9M \\
    & RainDropDiff$_{128}$~\cite{wd}            & 32.43 & 0.9334 &\hspace{0.1cm} \textgreater10  & 85.6M \\
    & Restormer~\cite{restormer}                & 31.99 & 0.9301 &\hspace{0.1cm} 0.21 & 26.5M \\
    & \dualline{DRS\red{f}ormer~\cite{drsformer}\\(w/o MEFC)} & 31.72 & 0.9342 &\hspace{0.1cm} 0.58 & 32.6M \\
    & I$^{2}$R$^{2}$Net~\cite{i2r2net}          & 28.74 & 0.9004 &\hspace{0.1cm} 0.55 & 2.1M  \\
    & \textbf{WeatherRemover (Ours)}             & \redbf{32.99} & \redbf{0.9435} &\hspace{0.1cm} 0.16 & 24.3M \\
    \hline
    \multirow{5}{*}{\centering \shortstack{Multi\\Weather}}
    & TransWeather~\cite{transweather}          & 28.17 & 0.8904 &\hspace{0.1cm} 0.01 & 38.1M \\
    & WeatherDiff$_{64}$~\cite{wd}              & 30.71 & \redbf{0.9312} &\hspace{0.1cm} \textgreater10  & 82.9M \\
    & WeatherDiff$_{128}$~\cite{wd}             & 29.66 & 0.9225 &\hspace{0.1cm} \textgreater10  & 85.6M \\
    & Restormer~\cite{restormer}                & 30.81 & 0.9261 &\hspace{0.1cm} 0.20 & 26.5M \\
    & \textbf{WeatherRemover (Ours)}             & \redbf{31.51} & 0.9307 &\hspace{0.1cm} 0.16 & 24.3M \\
    \hline
  \end{tabular}
  }
  \begin{flushleft}
    Note: The experiment is divided into two parts: single-weather raindrop removal and multi-weather raindrop removal. The text marked in bold red indicates the best PSNR or SSIM values in each part. Note that DRS\red{f}ormer requires MEFC disabled for training and evaluation on this dataset.
  \end{flushleft}
\vspace{-2.0em}
\end{table}

\textbf{Deraining and dehazing resluts.} As shown in Table~\ref{tb-odr}, we evaluated models trained on Outdoor-Rain~\cite{hrgan} and AllWeather~\cite{transweather} using a test set we separated from Outdoor-Rain. Similar to the results of raindrop removal, our model achieved state-of-the-art performance in single-weather scenarios. Additionally, our PSNR also is the highest in multi-weather scenarios.

\begin{figure*}[ht]
  \centering
  \includegraphics[scale=0.4]{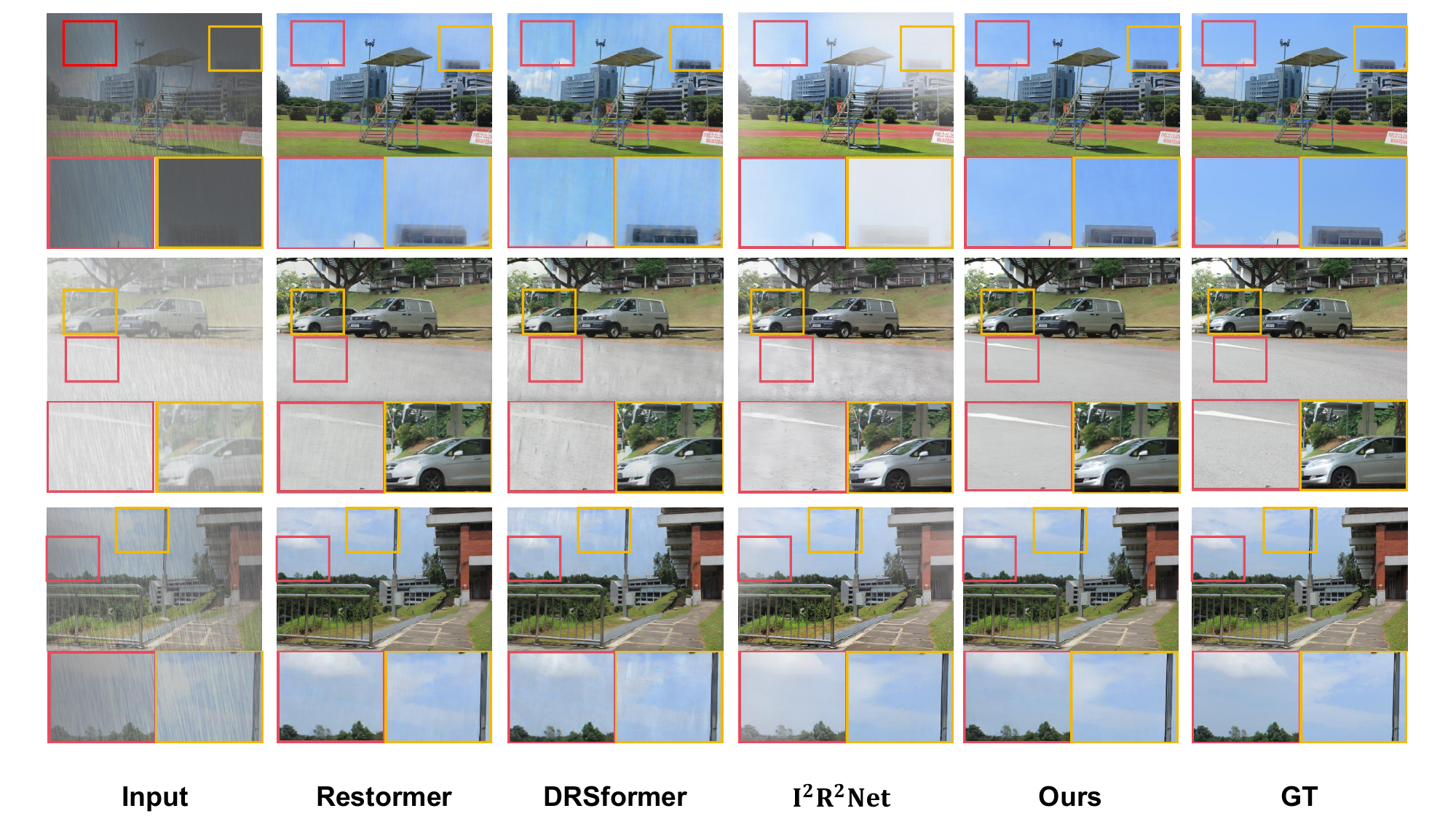}
  \caption{\red{\revise{Single-weather deraining and dehazing visual comparisons.}} From left to right, the images displayed are: \red{The original input images}, Restormer~\cite{restormer}, DRS\red{f}ormer~\cite{drsformer}, I$^2$R$^2$Net~\cite{i2r2net}, our proposed model, and the ground truth \red{images}. Below each image, magnified details are provided to emphasize the distinctions in performance. \rsix{The images are from the Outdoor-Rain~\cite{hrgan} dataset. The first and third rows demonstrate the removal of rain haze from the sky, while the second row compares the impact of different models on preserving the original image details.}}
  \label{fig:outdoor_rain}
\end{figure*}

In single-weather experiments, the most lightweight model is PCNet~\cite{pcnet}, which features the fewest parameters and the shortest inference time. However, its performance in restoring images affected by mixed rain and fog is notably inferior to Transformer-based models. In contrast, our model achieves approximately a 24\% improvement in PSNR and a 3\% improvement in SSIM, reaching the current state-of-the-art level. We present a comparative analysis of the performance of different models in eliminating rain and fog from images in Fig.~\ref{fig:outdoor_rain}. In scenarios involving sky backgrounds, the other three models struggle to effectively and thoroughly clear the fog, whereas our model delivers a notably superior outcome. Furthermore, when it comes to reconstructing background details, our model closely aligns with the ground truth, whereas the other models display varying levels of blurriness and loss of detail. 

\begin{figure*}[htbp]
  \centering
  \includegraphics[scale=0.4]{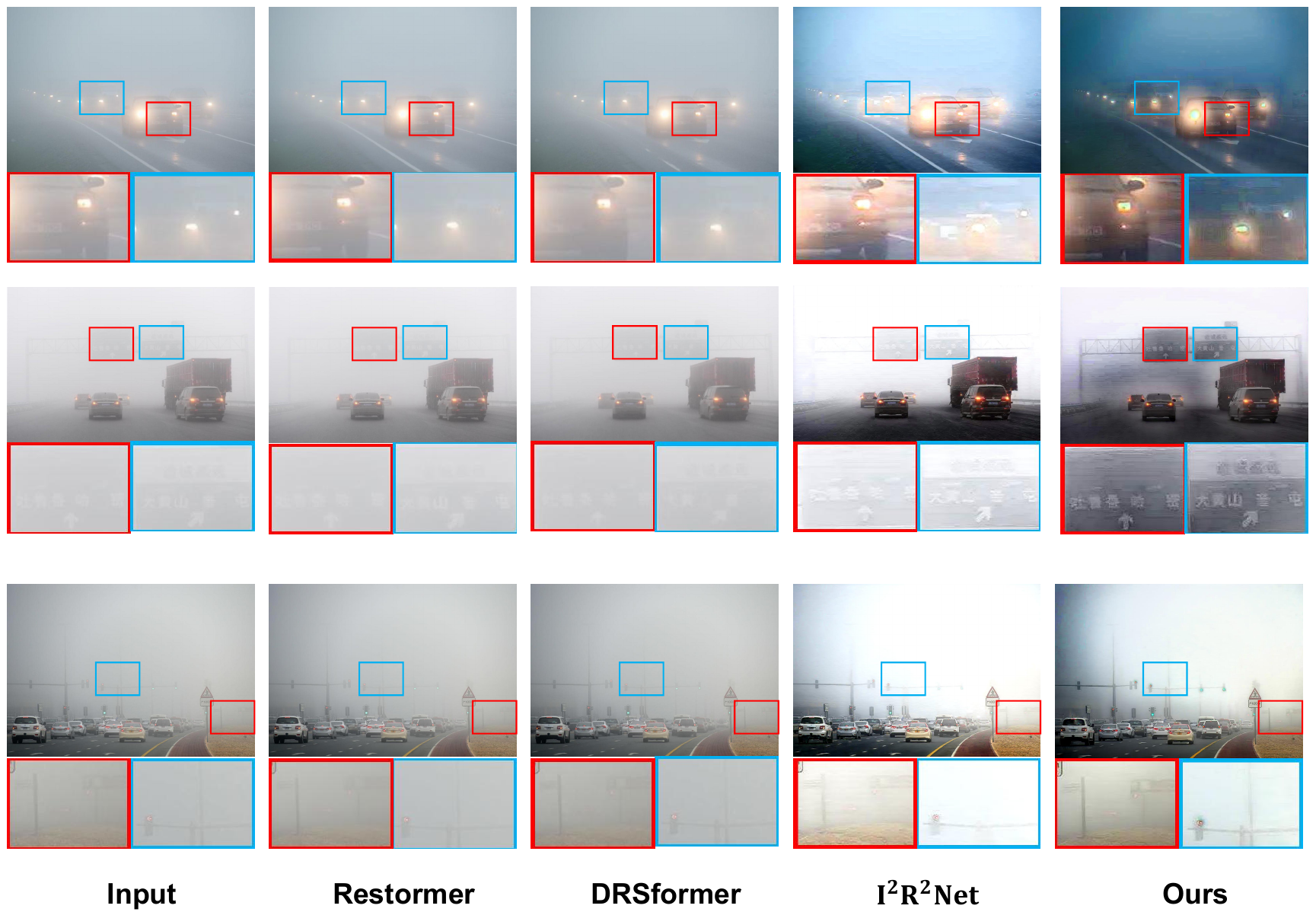}
  \caption{\red{\revise{Real-scene rain and dense fog removal visual comparisons.} From left to right: Input image, Restormer~\cite{restormer}, DRSformer~\cite{drsformer}, I$^{2}$R$^{2}$Net~\cite{i2r2net}, and our model. To show the fineness of the restoration, there are enlarged details below each image. \rsix{Images with extreme dense fog from the DAWN~\cite{dawn} dataset were selected to demonstrate the effectiveness of different models in fog removal and their capability to restore image details.}}}
  \label{fig:rain_fog}
\end{figure*}

\red{To thoroughly assess our model's effectiveness in extreme weather conditions, we conducted evaluations using real-world images of rain and dense fog. The results, illustrated in Fig.~\ref{fig:rain_fog}, demonstrate our model's capability to effectively restore images under these challenging conditions. \modify{The enlarged sections of the images in the first and second rows} highlight our model's proficiency in recovering details while preserving the original image information. \modify{Nevertheless, for images with more complex backgrounds, as shown in the third row, our model demonstrates limited dehazing effectiveness. It successfully removes nearby haze but struggles to restore intricate background details. This limitation stems from the model’s insufficient ability to interpret contextual information and accurately process haze-affected background features. Future improvements should prioritize enhancing the model’s capacity for context integration and refining its feature extraction and fusion mechanisms.}}

Under multiple weather conditions, although WeatherDiff$_{64}$~\cite{wd} achieved a slightly higher SSIM (0.0142 dB) than our model, it has a significantly longer inference time due to its generative model. While TransWeather has a 0.03-second shorter inference time, it falls short in other metrics. This demonstrates that our model also performs well in handling rain and fog under mixed-weather training conditions.

\red{The comparison tests indicate that our model strikes a favorable balance between performance and efficiency when compared to Restormer and DRSformer. Unlike TransWeather, our model offers greater flexibility in handling multiple weather conditions. Although WeatherDiffusion achieves commendable results in both single and multi-weather scenarios, it suffers from prolonged inference times due to its reliance on a diffusion model and the need to process images in segments~\cite{wd}. In contrast, our model is more lightweight and efficient. Furthermore, it demonstrates impressive detail restoration capabilities on both real and synthetic datasets, highlighting its potential and versatility across a wide range of application scenarios.}

\begin{table}[]
  \caption{Results of deraining and dehazing on the Outdoor-Rain dataset.}
  \centering
    \resizebox{\linewidth}{!}{
  \begin{tabular}{llllll}
    \hline
    \textbf{\hspace{0.15cm}Task} & Model & PSNR$\uparrow$ & SSIM$\uparrow$ & Time/s & Param \\
    \hline
    \multirow{10}{*}{\centering \shortstack{Deraining\\\&\\Dehazing}}
    & HRGAN~\cite{hrgan}                     & 21.56 & 0.8550 & \hspace{0.1cm}0.28 & 40.6M  \\ 
    & PCNet~\cite{pcnet}                     & 26.19 & 0.9015 & \hspace{0.1cm}0.02 & 0.6M   \\
    & MPRNet~\cite{mprnet}                   & 28.03 & 0.9192 & \hspace{0.1cm}0.02 & 3.6M   \\
    & RainHazeDiff$_{64}$~\cite{wd}          & 28.38 & 0.9320 & \hspace{0.1cm}\textgreater10  & 82.9M \\
    & RainHazeDiff$_{128}$~\cite{wd}         & 26.84 & 0.9152 & \hspace{0.1cm}\textgreater10  & 85.6M \\
    & Restormer~\cite{restormer}             & 31.28 & 0.9252 & \hspace{0.1cm}0.16 & 26.5M  \\
    & \dualline{DRS\red{f}ormer~\cite{drsformer}\\(w/o MEFC)} & 31.58& 0.9228 & \hspace{0.1cm}0.38 & 32.6M  \\
    & I$^{2}$R$^{2}$Net~\cite{i2r2net}       & 21.41 & 0.8738 & \hspace{0.1cm}0.42 & 2.1M   \\
    & \textbf{WeatherRemover (Ours)}          & \redbf{32.56} & \redbf{0.9330} & \hspace{0.1cm}0.13 & 24.3M \\
    \hline
    \multirow{5}{*}{\centering \shortstack{Multi\\Weather}}
    & TransWeather~\cite{transweather}       & 27.89 & 0.9179 & \hspace{0.1cm}0.01   & 38.1M \\
    & WeatherDiff$_{64}$~\cite{wd}           & 29.64 & \redbf{0.9312} & \hspace{0.1cm}\textgreater10  & 82.9M \\
    & WeatherDiff$_{128}$~\cite{wd}          & 29.72 & 0.9216 & \hspace{0.1cm}\textgreater10   & 85.6M   \\
    & Restormer~\cite{restormer}             & 30.53 & 0.9118 & \hspace{0.1cm}0.16  & 26.5M  \\
    & \textbf{WeatherRemover (Ours)}          & \redbf{31.52} & 0.9170 & \hspace{0.1cm}0.13 & 24.3M \\
    \hline
  \end{tabular}
  }
  \label{tb-odr}
  \begin{flushleft}
    Note: The experiment is divided into two parts: single-weather and multi-weather. The text marked in bold red indicates the best PSNR or SSIM values in each part. It is important to note that DRS\red{f}ormer necessitates the deactivation of MEFC for both training and evaluation on the Outdoor-Rain dataset.
  \end{flushleft}
  \vspace{-2.0em}
\end{table}

\subsection{Ablation Study}
Table~\ref{tb_gating} presents the results of our ablation study on the gating mechanisms used in our model's D-Stage and the FN of MS-PVT. It is important to note that during our evaluation of the inference times for the RainDrop and Outdoor-Rain datasets, we observed that the first image significantly impacted the average inference time. Therefore, we excluded the first image when calculating the average time. The findings indicate that in each dataset, both PSNR and SSIM are at their lowest without any gating mechanisms.

The gating mechanism in the FN, with approximately 3.66 M parameters, significantly impacts the model's performance more than the downsampling gating mechanism. In the RainDrop, Snow100K, and Outdoor-Rain datasets, the PSNR increased by 0.8615 dB, 0.4995 dB, and 1.2976 dB, respectively. Conversely, the downsampling gating mechanism, which has around 0.46 million parameters, shows a smaller PSNR improvement of 0.2526 dB, 0.0191 dB, and 0.0894 dB in each dataset. Notably, when both gating mechanisms are enabled, the model's overall performance improves significantly, with PSNR increases of 1.1133 dB, 0.7708 dB, and 2.0783 dB compared to the model without either mechanism in each dataset. This demonstrates the substantial positive impact of incorporating both gating mechanisms.

\rfive{Alongside the findings in Table~\ref{tb_gating}, we assessed the multiply-accumulate operations (MACs) in giga units (G) during inference on tensors with dimensions of $1 \times 3 \times 720 \times 480$, with gradients disabled. The lowest computational cost was observed when both the GFN and downsampling gating mechanisms were removed, yielding a MAC value of 280.12 G. Activating only the GFN led to an increase in MACs to 356.76 G, which is approximately 27\% higher than the configuration without any gating mechanisms. Conversely, enabling only the downsampling gating mechanism resulted in a modest 5\% increase in MACs. This indicates that while the GFN significantly boosts restoration performance, it also imposes a considerable computational load. When both gating mechanisms were activated, MACs reached their highest value of 377.21 G. Despite this peak in computational cost, the combined use of these mechanisms enhances the model's overall restoration capabilities to its fullest.}

\begin{table}[]
  \centering
  \caption{
  \revision{Ablation experiments of the gating mechanisms within GFN and D-Stage. Ablated by removing the element-wise multiplications and branches.
  }}
  \resizebox{\linewidth}{!}{
  \begin{tabular}{cclllll}
    \hline
    \multicolumn{1}{c}{} & \multicolumn{2}{c}{Gating} \\
    \cline{2-3} \vc{Dataset} & \revision{GFN} & D-Stage & \vc{PSNR$\uparrow$} & \vc{SSIM$\uparrow$} & \vc{Time/s} & \vc{Param}  \\
    \hline
                                   & \multicolumn{1}{l}{} &                                & 31.1566          & 0.9164          & 0.0805 & 20.10M \\
                                   & \checkmark           &                                & 32.0181          & 0.9250          & 0.0979 & 23.76M \\
                                   & \multicolumn{1}{l}{} & \multicolumn{1}{c}{\checkmark} & 31.4092          & 0.9184          & 0.0944 & 20.56M \\
    \multirow{-4}{*}{Snow100K}     & \checkmark           & \multicolumn{1}{c}{\checkmark} & \redbf{32.2699}  & \redbf{0.9268}  & 0.1161 & 24.32M \\
    \hline
                                   & \multicolumn{1}{l}{} &                                & 32.2264          & 0.9390          & 0.0967 & 20.10M \\
                                   & \checkmark           &                                & 32.7259          & 0.9432          & 0.1165 & 23.76M \\
                                   & \multicolumn{1}{l}{} & \multicolumn{1}{c}{\checkmark} & 32.2455          & 0.9390          & 0.1130 & 20.56M \\
    \multirow{-4}{*}{RainDrop}     & \checkmark           & \multicolumn{1}{c}{\checkmark} & \redbf{32.9972}  & \redbf{0.9435}  & 0.1352 & 24.32M \\
    \hline
                                   & \multicolumn{1}{l}{} &                                & 30.4891          & 0.9054          & 0.0962 & 20.10M \\
                                   & \checkmark           &                                & 31.7867          & 0.9231          & 0.1166 & 23.76M \\
                                   & \multicolumn{1}{l}{} & \multicolumn{1}{c}{\checkmark} & 30.5785          & 0.9045          & 0.1133 & 20.56M \\
    \multirow{-4}{*}{Outdoor-Rain} & \checkmark           & \multicolumn{1}{c}{\checkmark} & \redbf{32.5674}  & \redbf{0.9330}  & 0.1351 & 24.32M \\
    \hline
  \end{tabular}
  }
  \label{tb_gating}
  \begin{flushleft}
   \small
    Note: A checkmark denotes the activation of the gating mechanism within the module. Each model was trained and evaluated independently on the Snow100K, Raindrop, and Outdoor-Rain datasets. The values highlighted in red within each dataset correspond to the optimal performance metrics.
  \end{flushleft}
    \vspace{-2.0em}
\end{table}

\begin{table}[]
  \centering	 
  \caption{\revision{Ablation study results on different datasets using $1\times1$ convolution and depthwise separable convolution for generating query, key, and value in the MS-PVT.}}
\resizebox{\linewidth}{!}{
\begin{tabular}{cllllll}
\hline
\multicolumn{1}{l}{\hspace{0.3cm}Dataset} & \multicolumn{2}{l}{Method} & PSNR$\uparrow$ & SSIM$\uparrow$ & Time/s & Param \\ \hline
 & \multicolumn{2}{l}{1$\times$1Conv} & 30.3039 & 0.9075 & 0.1085 & 24.10M \\
 & \multicolumn{2}{l}{DWConv} & 31.8930 & 0.9230 & 0.1069 & 18.87M \\
 & \multicolumn{2}{l}{DWConv-1$\times$1Conv} & 31.8597 & 0.9223 & 0.1151 & 27.94M \\
\multirow{-4}{*}{Snow100K} & \multicolumn{2}{l}{1$\times$1Conv-DWConv} & {\color[HTML]{FF0000} \textbf{32.2699}} & {\color[HTML]{FF0000} \textbf{0.9268}} & 0.1161 & 24.32M \\ \hline
 & \multicolumn{2}{l}{1$\times$1Conv} & 32.5468 & 0.9418 & \multicolumn{1}{c}{0.1271} & 24.10M \\
 & \multicolumn{2}{l}{DWConv} & 32.6839 & 0.9431 & 0.1272 & 18.87M \\
 & \multicolumn{2}{l}{DWConv-1$\times$1Conv} & 32.7862 & 0.9433 & 0.1354 & 27.94M \\
\multirow{-4}{*}{RainDrop} & \multicolumn{2}{l}{1$\times$1Conv-DWConv} & {\color[HTML]{FF0000} \textbf{32.9972}} & {\color[HTML]{FF0000} \textbf{0.9435}} & 0.1352 & 24.32M \\ \hline
 & \multicolumn{2}{l}{1$\times$1Conv} & 29.2037 & 0.8907 & 0.1298 & 24.10M \\
 & \multicolumn{2}{l}{DWConv} & 31.0185 & 0.9135 & 0.1275 & 18.87M \\
 & \multicolumn{2}{l}{DWConv-1$\times$1Conv} & 31.0444 & 0.9150 & 0.1354 & 27.94M \\
\multirow{-4}{*}{Outdoor-Rain} & \multicolumn{2}{l}{1$\times$1Conv-DWConv} & {\color[HTML]{FF0000} \textbf{32.5674}} & {\color[HTML]{FF0000} \textbf{0.9330}} & 0.1351 & 24.32M \\ \hline
\end{tabular}
}
 \label{tb_conv}
  \begin{flushleft}
    Note: \revision{"DWConv" means depthwise separable convolution. Each dataset includes four experimental groups: the first two groups utilize only 1$\times$1 convolution or depthwise separable convolution, while the third and fourth groups swap the positions of the two convolutional layers.} The values highlighted in red indicate the best performance.
  \end{flushleft}
  \vspace{-2.0em}
\end{table}

Fig.~\ref{fig:pvtattn} illustrates the MS-PVT Attention mechanism, where the generation of the query is achieved through a 1$\times$1 convolutional layer followed by a \revision{depthwise separable convolution}. The same process is used for generating keys and values. To demonstrate the benefits of the 1$\times$1 convolutional layer and \revision{depthwise separable convolution} in MS-PVT Attention, we conducted ablation experiments on three datasets. The results are shown in Table~\ref{tb_conv}.

In each dataset, the first row in the table represents retaining only the 1$\times$1 convolution, the second row retains only the \revision{depthwise separable convolution}, and the third row swaps their positions. The results indicate that both modules enhance performance. Notably, \revision{depthwise separable convolution} has a more significant impact, increasing the PSNR by 1.5891 dB, 0.1371 dB, and 1.8148 dB across the datasets compared to the 1$\times$1 convolution. The inference times for using 1$\times$1 convolution and \revision{depthwise separable convolution} remain similar across all three datasets. However, \revision{depthwise separable convolution} results in 5.23 million fewer parameters than the 1$\times$1 convolution, demonstrating its effectiveness in the MS-PVT Attention mechanism. Despite this, the role of the linear transformation provided by the 1$\times$1 convolution should not be underestimated. As shown in the fourth row of the table for each dataset, when both the 1$\times$1 convolution and \revision{depthwise separable convolution} are used together, the PSNR increases by 0.3769 dB, 0.3133 dB, and 1.5489 dB, respectively, compared to using only the \revision{depthwise separable convolution}. Conversely, when their positions are swapped (as shown in the third row for each dataset), the PSNR and SSIM decrease and the number of parameters increases by 3.62 million, confirming the appropriateness of the original configuration.

\begin{table}[]
\centering
\caption{\modify{Ablation Study of Linear SRA and GFN on Different Datasets.}}
\resizebox{\linewidth}{!}{
\begin{tabular}{ccccccc}
\hline
Dataset & SRA & GFN & PSNR$\uparrow$ & SSIM$\uparrow$ & Time/s & Param \\ \hline
 &  &  & 31.7547 & 0.9232 & 0.0808 & 18.72M \\
 & \checkmark &  & 31.8923 & 0.9234 & 0.0962 & 20.56M \\
 &  & \checkmark & 32.1415 & 0.9267 & 0.1022 & 22.49M \\
\multirow{-4}{*}{Snow100K} & \checkmark & \checkmark & {\color[HTML]{FF0000} \textbf{32.2699}} & {\color[HTML]{FF0000} \textbf{0.9268}} & 0.1161 & 24.32M \\ \hline
 &  &  & 32.4960 & 0.9421 & 0.1027 & 18.72M \\
 & \checkmark &  & 32.6308 & 0.9409 & 0.1210 & 20.56M \\
 &  & \checkmark & 32.7250 & 0.9433 & 0.1298 & 22.49M \\
\multirow{-4}{*}{Raindrop} & \checkmark & \checkmark & {\color[HTML]{FF0000} \textbf{32.9972}} & {\color[HTML]{FF0000} \textbf{0.9435}} & 0.1352 & 24.32M \\ \hline
 &  &  & 30.8035 & 0.9204 & 0.0956 & 18.72M \\
 & \checkmark &  & 31.2700 & 0.9173 & 0.1137 & 20.56M \\
 &  & \checkmark & 32.0641 & 0.9285 & 0.1213 & 22.49M \\
\multirow{-4}{*}{Outdoor-Rain} & \checkmark & \checkmark & {\color[HTML]{FF0000} \textbf{32.5674}} & {\color[HTML]{FF0000} \textbf{0.9330}} & 0.1351 & 24.32M \\ \hline
\end{tabular}
}
\label{tb_sra_gfn}
  \begin{flushleft}
  \small
    \modify{Note: "\checkmark" indicates that the module is enabled. \redbf{The bold red font} highlights the optimal metrics.}
  \end{flushleft}
  \vspace{-2.0em}
\end{table}

\modify{To evaluate the contributions of the linear SRA and GFN modules within MS-PVT, Table~\ref{tb_sra_gfn} presents an ablation study across datasets in various scenarios. The results demonstrate that both modules significantly enhance the model’s restoration performance. Specifically, enabling linear SRA improves PSNR by 0.1376 dB, 0.1348 dB, and 0.4665 dB across the three scenarios, while GFN alone achieves even greater improvements of 0.3868 dB, 0.2290 dB, and 1.2606 dB, emphasizing its effectiveness in restoring fine details. Although GFN increases the parameter count by approximately 9\%, the performance gains justify the additional computational cost. When both linear SRA and GFN are combined, the model achieves optimal restoration performance across all datasets, highlighting the critical synergy between these modules in enhancing overall effectiveness.}

\subsection{Computation Analysis}

\begin{table}[htbp]
  \tabcolsep=0.2cm
  \centering
  \caption{Comparison of Transformer-based model inference computation on each dataset. Using MACs as a metric.}
  \begin{tabular}{lcc}
    \hline
    \multicolumn{1}{c}{} & \multicolumn{2}{c}{MACs/G} \\
    \cline{2-3} \multicolumn{1}{c}{\multirow{-2}{*}{Model}} & RainDrop-A & Snow100K-L \\
    \hline
    Restormer~\cite{restormer}            & 743.5        & 627.8 \\
    DRS\red{f}ormer~\cite{drsformer}            & 805.3        & 981.4 \\
    TransWeather~\cite{transweather}         & \redbf{32.6} & \redbf{28.7} \\
    LMQFormer~\cite{LMQFormer}            & 120.8        & 91.1 \\
    \textbf{WeatherRemover (Ours)} & 377.2        & 301.8 \\
    \hline
  \end{tabular}
  \label{tb_macs}
    \begin{flushleft}
    \small
    Note: DRS\red{f}ormer disables the MEFC to accommodate the RainDrop dataset. Since the image sizes of the RainDrop and Outdoor-Rain datasets are identical, their MACs metrics are the same. The best values are highlighted in red.
  \end{flushleft}
  \vspace{-2.0em}
\end{table}

The comparative experimental analysis indicates that Transformer-based models outperform both CNN and generative models in restoring images affected by multiple and single weather conditions. While WeatherDiffusion also produces favorable results, its reliance on a diffusion model results in significantly longer inference times and a larger number of parameters, which is its primary drawback. We evaluated the computational load of Transformer-based models through comparative experiments on three datasets: Snow100K-L, RainDrop-A, and Outdoor-Rain. The results, summarized in Table~\ref{tb_macs}, were measured using thop 2.0.6. \rfive{The evaluation metric was MACs, measured in giga units (G), representing billions of multiply-add operations per second.} A lower MACs value indicates a reduced computational load for the model. Notably, because the image sizes of RainDrop-A and Outdoor-Rain are identical, their MACs values are also the same.

Among the models analyzed, DRS\red{f}ormer, Restormer, and our model all employ CNN-based attention mechanisms. However, DRS\red{f}ormer has the highest computational cost due to its FN architecture~\cite{drsformer}. In contrast, our model uses a Transformer architecture with a linear SRA mechanism, resulting in significantly lower computational expenses. Specifically, our model achieves reductions of approximately 49\% and 53\% on the RainDrop dataset, and 51\% and 69\% on the Snow100K-L dataset, compared to Restormer and DRS\red{f}ormer, respectively. Furthermore, our model outperforms these models across all datasets. Our model also surpasses TransWeather in terms of parameter count and multi-weather restoration performance. However, the gating mechanism within our model’s FN increases its computational load, making it approximately ten times higher than that of TransWeather, necessitating further optimization. On the other hand, LMQFormer features a minimal number of parameters and fast inference time, but is limited to snow removal tasks, restricting its broader applicability.

\modify{Based on the earlier computational analysis, the WeatherRemover model demonstrates higher MACs than TransWeather and LMQFormer. However, it achieves about a 50\% reduction in MACs compared to Restormer and DRSformer, which share similar architectures. This indicates that the WeatherRemover model offers superior computational efficiency relative to these two models, highlighting its relatively lightweight design in terms of computational load.}

\subsection{Memory Analysis}

\begin{table}[t]
\tabcolsep=0.2cm
  \centering
  \caption{Comparison of Transformer-based model video memory consumption, measured in MiB.}
\begin{tabular}{lcc}
\hline
Model & \multicolumn{1}{l}{Memory (MiB)} & \multicolumn{1}{l}{Param} \\ \hline
\begin{tabular}[c]{@{}l@{}}DRSformer~\cite{drsformer}\\ (w/o MEFC)\end{tabular} & 140.9 & 32.6M \\
DRSformer~\cite{drsformer} & 147.8 & 33.6M \\
Restormer~\cite{restormer} & 114.2 & 26.5M \\
TransWeather~\cite{transweather} & 161.6 & 38.1M \\
LMQFormer~\cite{LMQFormer} & {\color[HTML]{FF0000} \textbf{27.1}} & {\color[HTML]{FF0000} \textbf{2.2M}} \\
\textbf{WeatherRemover (Ours)} & 109.6 & 24.3M \\ \hline
\end{tabular}
\label{tb_mem}
\begin{flushleft}
    \small
    Note: All models are set up for evaluation mode with gradients disabled for inference. The dimension of the inference tensor is 1$\times$3$\times$640$\times$480. The best values are highlighted in red.
  \end{flushleft}
  \vspace{-2.0em}
\end{table}

\red{
The video memory overhead of a model is a crucial factor in assessing its performance and directly influences its suitability for deployment. Table~\ref{tb_mem} provides a comparison of the memory usage of various Transformer-based models included in the study, along with their respective parameter counts. In this evaluation, all models were set to evaluation mode, and gradient accumulation was disabled to accurately measure the memory required for a tensor with inference dimensions of 1$\times$3$\times$640$\times$480 in MiB.
}

\red{
The measurement results reveal that LMQFormer stands out with the lowest memory usage and parameter count among all the models evaluated. This underscores its efficiency in terms of resource consumption and complexity. However, its dependence on pre-defined masks makes its performance heavily reliant on the accuracy of these masks~\cite{LMQFormer}. Moreover, the model's adaptability to different scenarios is limited, which restricts its broader applicability. On the other hand, DRSFormer and TransWeather possess a high number of parameters and significant memory overhead, indicating that they require substantial computational resources to function effectively. In contrast, both Restomer and our model exhibit a lower memory overhead and a moderate parameter count, reflecting a more optimal balance between performance and complexity.
}

\modify{Based on the GPU memory usage comparison in Table~\ref{tb_mem}, LMQFormer exhibits the lowest GPU memory consumption and parameter count, but it is specifically designed for snow removal. Among models capable of addressing multiple weather conditions, WeatherRemover achieves the optimal balance, delivering minimal GPU memory usage and parameter count while maintaining superior restoration performance. This underscores its lightweight design and effectiveness in handling diverse weather scenarios.}

\subsection{Inference Time Analysis}

\begin{table}[]
\centering
  \caption{\modify{Inference Time Comparison of Transformer-Based Multi-Weather Models.}}
  \resizebox{\linewidth}{!}{
\begin{tabular}{ccccccc}
\hline
 & \multicolumn{2}{c}{Snow100K} & \multicolumn{2}{c}{RainDrop} & \multicolumn{2}{c}{Outdoor-Rain} \\ \cline{2-7} 
\multirow{-2}{*}{Model} & T4 & 4090 & T4 & 4090 & T4 & 4090 \\ \hline
Restormer~\cite{restormer} & 0.66 & 0.07 & 0.99 & 0.11 & 0.78 & 0.08 \\
TransWeather~\cite{transweather} & {\color[HTML]{FE0000} \textbf{0.38}} & {\color[HTML]{FE0000} \textbf{0.04}} & {\color[HTML]{FE0000} \textbf{0.05}} & {\color[HTML]{FE0000} \textbf{0.01}} & {\color[HTML]{FE0000} \textbf{0.05}} & {\color[HTML]{FE0000} \textbf{0.00}} \\
DRSformer~\cite{drsformer} & 1.06 & 0.11 & 2.73 & 0.30 & 0.77 & 0.02 \\
\textbf{WeatherRemover (Ours)} & 0.39 & 0.05 & 0.66 & 0.07 & 0.54 & 0.05 \\ \hline
\end{tabular}
}
\begin{flushleft}
    \small
    \modify{Note: The employed devices are \textbf{NVIDIA TESLA T4} and \textbf{NVIDIA RTX 4090}, with inference conducted three times per dataset and averaged. \redbf{The bold red font} indicates the shortest inference time, and the unit of measurement is seconds.}
  \end{flushleft}
  \label{tb_time}
  \vspace{-2.0em}
\end{table}

\modify{To further evaluate the model’s inference time, Table~\ref{tb_time} provides a reliable assessment. We tested the transformer-based multi-weather models on NVIDIA TESLA T4 and NVIDIA RTX 4090, conducting three inference iterations per dataset and averaging the results.}

\modify{The findings reveal that TransWeather achieves the fastest inference time, particularly on the RainDrop dataset, where it outperforms our model by 92\% and 86\% on the two devices, respectively. While our model ranks second in inference speed, TransWeather suffers from limitations in restoration capability. In contrast, our model maintains efficient computation across various datasets and hardware environments while delivering superior restoration performance. This underscores its ability to balance computational efficiency and restoration quality, emphasizing its practical value.}

\subsection{Scaling with Resolution}
\begin{figure}[]
  \centering
  \includegraphics[scale=0.25]{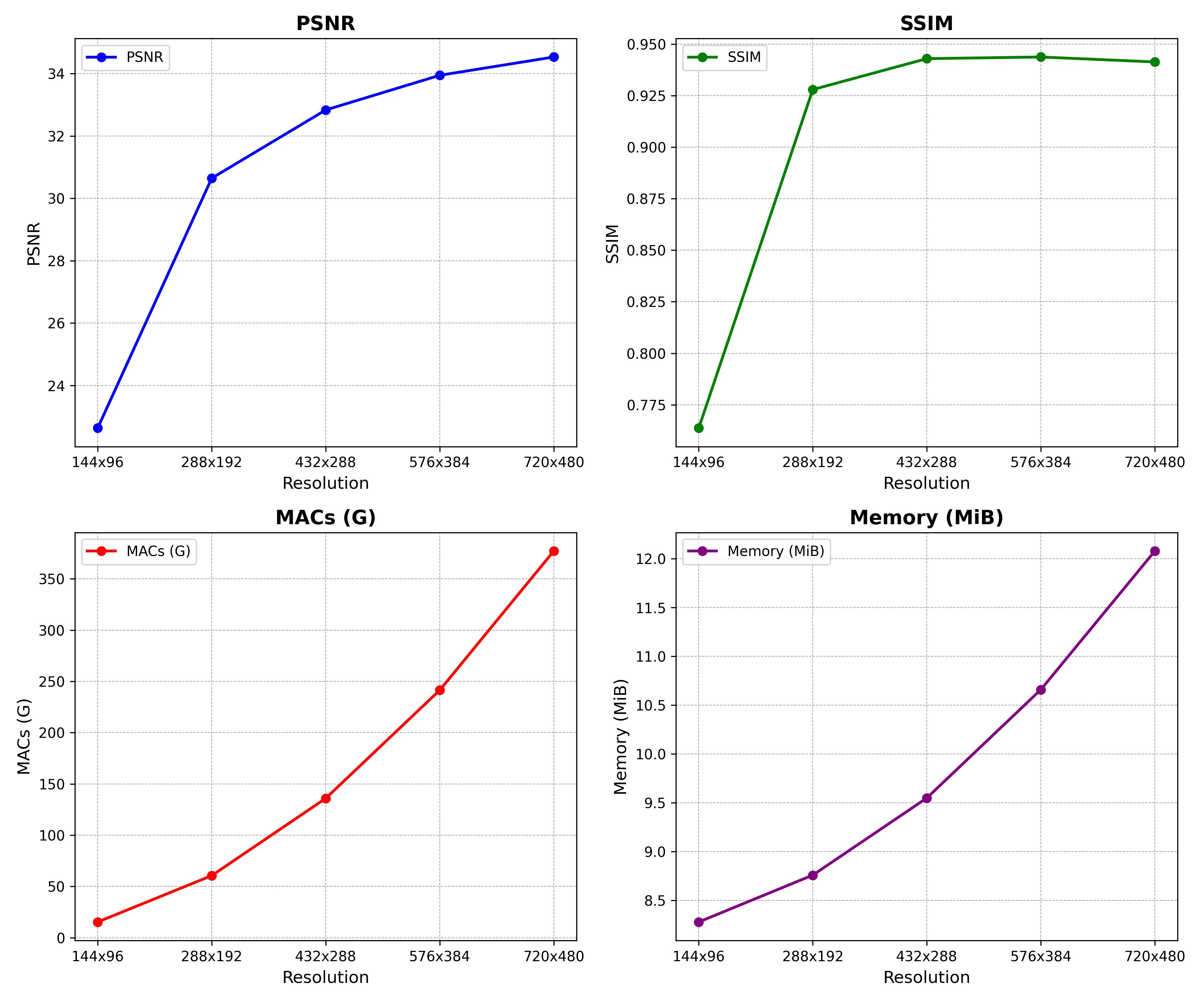}
  \caption{\rfive{Model performance vs. resolution. Select raindrops as an example to measure memory usage during inference while disabling gradients.}}
  \label{fig:resolution}
\end{figure}

\rfive{The model's restoration performance and computational costs fluctuate with different image resolutions. To examine how the model's effectiveness and computational efficiency change with resolution, Fig.~\ref{fig:resolution} illustrates the curves depicting variations in PSNR, SSIM, MACs, and memory usage across different resolutions.}

\rfive{As image resolution increases, there is a significant improvement in PSNR and SSIM values, reflecting enhanced restoration quality by the model. Essentially, higher resolutions yield better restoration outcomes. However, this improvement comes with a considerable rise in computational costs. Both MACs and memory usage increase sharply with higher resolutions, indicating that while the model delivers superior performance for high-resolution images, it also demands substantial computational resources.}

\section {Limitations and Future Works}
\revision{\revise{
Our model demonstrates strong performance in restoring images affected by both single and multiple weather conditions, striking a balance between efficiency and effectiveness. However, it does have certain limitations. Compared to existing models, our model falls short of TransWeather~\cite{transweather} and LMQFormer~\cite{LMQFormer} regarding inference time and computational load, highlighting deficiencies in computational and space utilization efficiency. Additionally, our model's performance in snow removal tasks lags behind that of Restormer~\cite{restormer}, indicating that its effectiveness is not consistently balanced across different types of tasks; it performs better for rain-related tasks. Furthermore, in multi-weather removal tasks, there remains a significant gap in our model's restoration efficacy across all scenarios compared to its performance under each single weather condition. These limitations are primarily attributed to the following challenges.}}

\begin{table}[]
\tabcolsep=0.10cm
  \centering
  \caption{\modify{Performance comparison of different feed-forward networks.}}
\begin{tabular}{cccc}
\hline
Model & MACs/M & Memory/MiB & Param \\ \hline
Restormer~\cite{restormer} & 58.06 & 3.59 & 189 \\
DRSformer~\cite{drsformer} & 318.27 & 3.59 & 1036 \\
TransWeather-E~\cite{transweather} & {\color[HTML]{FE0000} \textbf{27.65}} & 11.64 & {\color[HTML]{FE0000} \textbf{105}} \\
TransWeather-D~\cite{transweather} & 55.3 & 11.64 & 207 \\
\textbf{WeatherRemover (Ours)} & 49.77 & {\color[HTML]{FE0000} \textbf{3.52}} & 189 \\ \hline
\end{tabular}
\label{tb_gfn}
\begin{flushleft}
    \small
    \modify{Note: The MACs and GPU memory usage metrics are obtained by performing inference on a tensor of dimension 1$\times$3$\times$640$\times$480 with gradients disabled. TransWeather-E and TransWeather-D represent the feed-forward network configurations utilised in the encoder and decoder stages, respectively.}
  \end{flushleft}
  \vspace{-2.0em}
\end{table}

\begin{itemize}
\item
\rsix{
\textbf{High Computational Demand of The GFN.} \modify{Table~\ref{tb_gfn} presents a comparison of feed-forward network performance across various Transformer-based models designed for multi-weather support. The WeatherRemover model exhibits marginally better GPU memory efficiency, reducing memory usage by 0.07 MiB compared to Restormer and DRSformer. However, its MACs and parameter count are approximately 44\% higher than those of TransWeather-E. This increase stems from the GFN’s use of two identical branches to process feature maps across all channels, leading to greater computational overhead and parameter complexity.}
}
\item
\rsix{
\textbf{Difficulty in Handling Large Weather Occlusions.} The visual results in Fig.~\ref{fig:snow_real} and Fig.~\ref{fig:rain_fog} show that our model performs inadequately in handling dense occlusions and low-resolution restoration during extreme weather conditions, such as heavy fog and snow. This indicates limitations in the attention mechanism's ability to capture distant information effectively. 
}
\item
\rsix{
\textbf{Inefficient Multi-Weather Expansion.} Our model produced satisfactory results in restoring images affected by multiple weather conditions. \modify{However, restoration performance in multi-weather conditions remains inferior to that in single-weather scenarios. Enhancing capabilities necessitates incorporating new weather data into the existing dataset and retraining the model from scratch. Otherwise, training exclusively on new data risks catastrophic forgetting~\cite{learn_forget}, diminishing the model’s effectiveness in previously learned conditions.}
}
\end{itemize}

\revision{To address the aforementioned limitations, we intend to explore and implement several optimization techniques in future work.}
\begin{itemize}
    \item \rsix{\textbf{Optimization of Computational Efficiency and Memory Utilization}. The current dual-branch design of the GFN results in considerable computational and memory demands. To address this issue, we intend to substitute the convolutional layers in each branch with average and max pooling layers. This adjustment will allow each branch to handle specific feature map channels instead of processing all channels, decreasing computational requirements while maintaining thorough feature integration.}
    \item \rsix{\textbf{Enhanced Contextual Feature Integration.} To improve the model's capability in restoring images impacted by severe weather and significant obstructions, we propose investigating a global attention mechanism. This method is intended to effectively incorporate contextual features, thereby increasing the model's overall performance.
    \rsix{Additionally, we aim to adapt this mechanism to dynamic situations, including changing weather conditions and varying resolutions. This adaptation is expected to improve the model’s proficiency in restoring low-resolution images, exemplified by removing snow from blurry images, ultimately broadening the model's applicability across diverse scenarios.}}
    \item \rsix{\textbf{Continual Learning.} \modify{The task of multi-weather removal can be perceived as an incremental learning challenge, wherein the model progressively improves its capability to manage new weather conditions while maintaining its performance with existing ones. To tackle the imbalance present in real-world datasets, the AIR method~\cite{air} can be utilized to compute reweighting factors, ensuring that all weather categories contribute equally to the loss function. Instead of retraining the entire model, adaptation to novel weather conditions can be accomplished by freezing the encoder and updating only the decoder's parameters with new data.} These strategies are designed to mitigate issues related to forgetting in incremental learning and to handle imbalanced data distribution, thereby allowing effective restoration of images across a wider spectrum of weather scenarios.}
\end{itemize}

\revision{Beyond the optimizations mentioned previously, we are also planning to explore real-time applications in future work. \rfive{According to the inference times reported in comparative experiments (Tables~\ref{bk-s100k}, ~\ref{tb-rd}, and~\ref{tb-odr}), the model achieves frame rates of 9.09 FPS for snow removal, 6.25 FPS for raindrop removal, and 7.69 FPS for rain-fog removal in single-weather scenarios, resulting in an average frame rate of 7.68 FPS for video applications. However, as highlighted in Table~\ref{tb_gfn}, the GFN structure within the MS-PVT module contributes significant computational overhead. To enhance the frame rate, the gating mechanism would need to be deactivated, though this adjustment may compromise restoration quality.} Currently, this frame rate is suitable for applications such as outdoor system monitoring in static scenes, where it can enhance the accuracy of object recognition under adverse weather conditions. Once we improve computational efficiency, our model will be capable of supporting higher frame rates, making it applicable for dynamic scenarios, such as in cameras used for autonomous driving.}

\section{Conclusion}

\red{\rsix{In this paper, we present WeatherRemover, a model specifically developed to eliminate weather-related distortions from images, either caused by single or multiple factors.} The model employs a Transformer architecture enhanced with linear spatial reduction and channel-wise attention mechanisms based on convolution within a gated down-sampling UNet framework. This strategic design lowers the model's computational requirements, reduces inference time, and enhances the focus on key image features. Additionally, it preserves consistent dimensions between input and output, allowing for precise restoration by emphasizing local feature details. \revision{Experimental results demonstrate that WeatherRemover effectively removes snow, raindrops, and combinations of rain and fog across various weather conditions. It achieves a commendable balance between restoration effectiveness and computational efficiency. Recognizing WeatherRemover's success in restoring realistic scenes, particularly those affected by rain and fog, it can significantly enhance the accuracy of subsequent tasks, such as semantic segmentation and object tracking in challenging weather conditions. By improving the quality of degraded images, WeatherRemover offers a novel solution for outdoor surveillance and automated driving assistance systems.}}

{
\bibliographystyle{IEEEtran}
\bibliography{IEEEabrv,references}
}

\end{document}